\documentclass[12 pt]{article}
\usepackage[left=1in,right=1in,top=1.2in,bottom=1.2in,
            footskip=.25in]{geometry}
\usepackage{amsmath, amssymb,amstext}
\usepackage{subcaption}
\usepackage{graphicx} % Required for inserting images
\usepackage{amsmath}
\usepackage{authblk}
\usepackage{color}
\usepackage{algorithm}
\usepackage{algorithmic}
\usepackage{amssymb}
\usepackage{natbib}
\usepackage{multirow}
\usepackage{diagbox}
\usepackage{url}
\usepackage{titlesec}
\usepackage{hyperref}

\title{Learning Coupled System Dynamics under Incomplete Physical Constraints and Missing Data}
\author[1]{Esha Saha}
\author[1]{Hao Wang}

\affil[1]{Department of Mathematical and Statistical Sciences, University of Alberta, Canada}
\date{}

\begin{document}

\maketitle
% \tableofcontents
\begin{abstract}
    Advances in data acquisition and computational methods have accelerated the use of differential equation based modeling for complex systems. 
    % Modeling complex systems governed by coupled partial differential equations is challenging when physical laws are known for only a subset of variables and the remaining variables are accessible only through data.
Such systems are often described by coupled (or more) variables, yet governing equation is typically available for one variable, while the remaining variable can be accessed only through data. 
This mismatch between known physics and observed data poses a fundamental challenge for existing physics-informed machine learning approaches, which generally assume either complete knowledge of the governing equations or full data availability across all variables.
In this paper, we introduce MUSIC (\textbf{M}ultitask Learning \textbf{U}nder \textbf{S}parse and \textbf{I}ncomplete \textbf{C}onstraints), 
% the General Regularized Architecture for Sparse Physics (MUSIC),
a sparsity induced multitask neural network framework that integrates partial physical constraints with data-driven learning to recover full-dimensional solutions of coupled systems when physics-constrained and data-informed variables are mutually exclusive. 
MUSIC employs mesh-free (random) sampling of training data and sparsity regularization, yielding highly compressed models with improved training and evaluation efficiency.
We demonstrate that MUSIC accurately learns solutions (shock wave solutions, discontinuous solutions, pattern formation solutions)  to complex coupled systems under data-scarce and noisy conditions, consistently outperforming non-sparse formulations. 
These results highlight MUSIC as a flexible and effective approach for modeling partially observed systems with incomplete physical knowledge.
\end{abstract}

\section{Introduction}
Mathematical modeling and simulation forms the core of many problems in science and engineering with applications in biological patterns of organisms, forecasting various environmental phenomenon, or simulating the energy transfer between two bodies, etc. 
One of the most common techniques of modeling involves using \textit{rates of change} to build systems of differential equations, either based on ordinary differential equations (ODEs) and/or partial differential equations (PDEs), often derived from first principles such as the conservation laws or experimental or knowledge-based derivations.
However, for many complex systems the full set of governing equations are often ambiguous or partially unknown since physical interactions between variables are undiscovered and ill-defined \citep{chen2021physics}.
Even for systems with well-understood and defined governing equations, analytic solutions are often intractable, leading to widespread reliance on numerical methods such as finite difference, finite volume, and finite element methods.
However, numerical methods suffer from the curse of dimensionality and often require substantial computational resources, as fine spatial and/or temporal discretizations are needed to ensure numerical stability for most schemes. 
As system complexity increases, the number of grid points grows exponentially, particularly for higher-order systems that more accurately represent real-world phenomena.
In parallel, advances in data acquisition technologies have enabled data-driven approaches for discovering governing equations and solving complex systems, significantly transforming modern modeling and simulation practices across science and engineering disciplines.
For example, using observational data of methane concentrations (e.g. weather monitoring stations, satellite imagery, etc) near methane emitting sources, we can train a model to help uncover true emissions rates from sources whose underlying methane emitting dynamics are unknown \citep{saha2025dispersion}.
Advances in machine learning theories, computational capacity, and data
availability have introduced a fresh perspective towards data-driven solving and discovery of governing equations \citep{raissi2019physics,chen2021physics,luo2025physics,stephany2024pde,berg2019data,saha2023diffusion,jacot2018neural,rudy2017data,saha2023spade4,brunton2016discovering}.

The use of additional information to train a machine learning model is not new to the community. Earlier works have established the usefulness of adding side information to in learning dynamical systems \citep{ahmadi2020learning,greydanus2019hamiltonian,kang2017incorporating}.
In a pioneering work on identifying governing equations of a system of ODEs from data, \citep{brunton2016discovering} proposed sparse identification of nonlinear dynamics
(SINDy), which selected dominant candidate functions from a high-dimensional dictionary of candidate functions using sparse regression. 
In the past few years, SINDy has captured the interest of many research groups, leading to identifying nonlinear dynamical systems in various applications, such as fluid flows \citep{sato2025rheo,foster2022estimating}, epidemic dynamics \citep{jiang2021modeling,horrocks2020algorithmic,pani2025novel}, atmospheric sciences \citep{yang2024atmospheric,guo2024uncertainty,teruya2024data}, wildfire modeling \citep{greenwood2024data}, predator–prey systems \citep{dam2017sparse,frank2024evaluating}, to name a few. 
The sparsity-promoting paradigm
has also been extended for the data-driven discovery of spatiotemporal systems governed by PDEs \citep{rudy2017data}, where the dictionary is built by incorporating spatial partial derivative terms. 
Despite its wide applicability, the SINDy framework suffers from dependence on quality and quantity of the measured
data and errors introduced by numerical differentiation. 
To avoid the issues of numerical derivatives and harness automatic differentiation, \citep{raissi2019physics} proposed physics-informed neural networks (PINNs) in a pioneering attempt to solve nonlinear PDEs using deep neural networks. 
The proposed framework learns by optimizing the parameters based on a loss function defined on available data and physical constraints.  
There has been an exponentially large number of works that apply PINNs for
tackling a wide range of scientific problems in fields of physics \citep{cai2021physics}, biology \citep{lagergren2020biologically,ahmadi2024ai}, ecology \citep{chrisnanto2025unified,wesselkamp2024process,saha2025dispersion}, etc when the explicit form of PDEs is known.
Some other works \citep{chen2021physics} have also extended the PINN formulation to learn the governing equations from a dictionary of candidate terms created from the outputs of the DNN using sparse regression.
While all these works have shown remarkable accuracy in learning solutions of the systems and/or governing equations, they all have an underlying assumption that limits their applicability.
All the above mentioned formulations for high dimensional systems assume that either data or the physics is known for all the variables in the system.
This is however untrue for most real-world applications.
For example, consider the 2D Reaction-Diffusion system used to model wildfire spread with solutions $(u,\beta)$ where $u$ denotes the fire-front temperature and $\beta$ denotes the fuel availability.
In order to solve this system, one first needs to discern the underlying system.
While the vector field of $\beta$ is often known from a vast literature modeling fuel availability/burning \citep{anand2017physics,mell2007physics}, collecting real time on-field fuel availability data is challenging (fuel availability is driven by multiple factors such as humidity, vegetation, etc).
On the other hand, modeling the vector field for $u$ is non-trivial due various environmental interactions that may be unknown. 
However, observational data for $u$ may be available through various sources such as satellite imagery. 
In such a scenario when the available data and known physics are mutually exclusive, all the proposed formulations for obtaining a full dimensional coupled solution $(u,\beta)$ fail.

In order to fill this gap in literature, we propose the Multitask Learning Under Sparse
and Incomplete Constraints (MUSIC), a novel mesh-free training framework of sparse multitask neural network  that learns solutions to coupled PDEs when the available data and known physics are mutually exclusive. 
Our approach incorporates the power of DNNs for a rich representation learning with $\ell_0$ sparsity enforced on the DNN to induce model compression in order to tackle the limitations of existing methods that scale poorly with computational resources, data noise and scarcity. 
The mesh-free sampling frees the model limitations driven by uniform sampling of spatio-temporal data for trainnig.
In particular, our major contributions are highlighted below.
\begin{itemize}
    \item We propose sparse multitask neural network with physics informed solution learning for coupled PDEs when the available data and known physics are mutually exclusive.
    Other than improved errors for scarce and noisy data, model sparsity offers several advantages which include lower training time, lesser computational complexity for model evaluation
    \item We use a mesh-free (random) sampling to create the training dataset, omitting the need for uniform data sampling, making it highly useful for real-world applications where uniform data sampling is not possible.
    Moreover, the multitask DNN uses the training data for enforcing both data fitting as well as physical constraints, leading to the omission of separate collocation points.
    
    % \item For noisy data, this model is especially useful since we only use data from one of the variables, thus effectively mitigating the effects of noise from the other variable.
\end{itemize}

\section{Model Framework}\label{sec:model-framework}
To motivate this work, we recall the example stated in the previous section.
Consider the 2D Reaction-Diffusion system used to model wildfire spread with solutions $(u,\beta)$ where $u$ denotes the fire-front temperature and $\beta$ denotes the fuel availability.
Solving this system (numerically or analytically) can be challenging due to various reasons such as incomplete modeling (unknown dynamics between variables), presence of discontinuous source term(s), unknown environmental variable interactions, etc.
When data for only the fire-front temperature $u$ is available, data-driven methods prove to be inadequate without data for $\beta$, which is hard to collect in real time. 
While the vector field of $\beta$ is often known from a vast literature modeling fuel availability/burning \citep{anand2017physics,mell2007physics}, physics constrained ML methods \citep{raissi2019physics,chen2021physics} are inapplicable without a known vector field for $u$ (which is much harder to model due various environmental interactions), since these formulations assume full dimensional physics (or data, or both).
In such a scenario, we want to explore the possibility of solving the system to obtain a full-field solution of the system. 

More formally, let $\Omega\subset\mathbb{R}^d$ be a spatial domain and $T>0$ be a final time. Our goal is to learn the solutions $\mathbf{u} = \{u_1,u_2,\dots,u_n\}$, $\mathbf{u}:\Omega\times (0,T]\rightarrow\mathbb{R}^n$ governed by a set of $n$ partial differential equations (PDEs) i.e.,
\begin{equation}
    \begin{aligned}
        \dfrac{\partial u_i}{\partial t} &= F_i(u_1,\dots,u_n,\nabla u_1,\dots,\nabla u_n,\nabla^2 u_1,\dots.\nabla^2 u_n,\dots),\,\, i=1,\dots,n;\\
        u_i(\mathbf{x},0) &= u_i^0(\mathbf{x}), \,\, \mathbf{x}\in\Omega;\\
        u_i(\mathbf{x},t) & = g_i(\mathbf{x},t),\,\, \mathbf{x}\in\partial\Omega, t\in(0,T],
    \end{aligned}
\end{equation}
where $\nabla$ denotes spatial derivatives and $\partial\Omega$ denotes the boundary of the domain. 
In this paper, our goal is to learn the full field $\mathbf{u}$ with the key assumption of the unavailability of full dimensional data or physics. 
In previous works, the authors assume that either the physics of the full system is known, or data for all the variables are available, or both. 
This is often unrealistic for most applications, where it might be possible to model some variables using equations but not all. Referring to such solutions as the \textit{equation variable}, collecting data for such variables might not be feasible for verifying the modeled equations.
On the other hand, it may be possible to measure some variables, hereby referred to as the \textit{data variable}, the physical laws may not be obvious or hard to model.
% Moreover, data for $u_i$'s are also scare i.e, very small in comparison to the entire domain.
Additionally, the challenge of learning solutions from mutually exclusive priors often gets paired with data scarce regimes, incomplete physical models and noisy training data.
Inspired from learning solutions to PDEs using the a physics based regularization, we propose a sparse multitask model optimized using a loss function jointly defined both on the data fitting and the known equations.
The model learns the full-field solution variables simultaneously using shared model parameters which are used parallelly to define the physics-based regularization.
% Our sparse multitask model MUSIC, learns the solutions with shared parameters that help learn the dependency of the variables on each other. 
For $0<k<n$, suppose measured data for solutions $\{u_1,\dots,u_k\}$ are available (\textit{data variables}) and the functions $\{F_{k+1},\dots,F_n\}$ (vector fields of the \textit{equation variables}) are known such that there is no overlap between the equation and data variables.
Our goal is to learn the full solution $\{u_1,\dots,u_k,\dots,u_n\}$ in a given time interval.
The sparsity constraints on the model speeds up training thereby reducing computational resources. The final trained model is a highly sparse, thus reducing storage and evaluation costs.
% We first use a fully connected neural network to learn the solutions.
% Based on available priors, we then regularize the loss functions accordingly to learn the sparse weights of the model.

Let $X = \{(\mathbf{x}_i,t_i)\}_{i=1}^m$ denote $m$ spatiotemporal points in the domain for which data is available i.e., $U=\{u_1^i,\cdots,u_k^i\}_{i=1}^m$ denote data matrix of the set of data variables.
Let $\widehat{U} = \{\widehat{u}^i_1,\cdots,\widehat{u}^i_k,\cdots,\widehat{u}_n\}_{i=1}^m$ denote the outputs of a $L$ layer surrogate (neural network) model with parameters $\Theta$ such that
\begin{equation}
    \widehat{U} = \Theta_L\cdots\phi(\Theta_2\phi(\Theta_1 X))
\end{equation}
where $X\in\mathbb{R}^{(d+1)\times m}$ denotes the input data matrix, $\widehat{U}\in\mathbb{R}^{n\times m}$ denotes the full-field learned solutions of the system, $\Theta_1 \in\mathbb{R}^{H\times (d+1)}$, $\Theta_2,\cdots\theta_{L-1} \in\mathbb{R}^{H\times H}$, $\Theta_L \in\mathbb{R}^{n\times H}$ denotes the weights (and biases) of the neural network and $\phi$ denotes the activation function. 
Suppose $\{\widehat{F}_{k+1}(\widehat{u}^i_1,\cdots,\widehat{u}^i_n),\cdots,\widehat{F}_{n}(\widehat{u}^i_1,\cdots,\widehat{u}^i_n)\}_{i=1}^m$ denotes the known physical equations evaluated at the predicted outputs. Then the loss is given by
% \begin{equation}
\begin{align}
    L &= \lambda_1\|u_1 - \widehat{u}_1\|_2^2 + \cdots +\lambda_k\|u_k - \widehat{u}_k\|_2^2 \label{eq:L_data}\\
    & + \lambda_{k+1}\left\|\dfrac{\partial \widehat{u}_{k+1}}{\partial t} - F_{k+1}(\widehat{u}_1,\cdots,\widehat{u}_n) \right\|_2^2 +\cdots +\lambda_n\left\|\dfrac{\partial \widehat{u}_n}{\partial t} - F_n(\widehat{u}_1,\cdots,\widehat{u}_n) \right\|_2^2 \label{eq:L_phy}\\
    &+ \Lambda\|\mathbf{\Theta}\|_0 \label{eq:L_l0}
    \end{align}
% \end{equation}
where $\{\lambda_i\}_{i=1}^n, \Lambda$ are the penalty parameters that balance the trade-offs between data fitting, PDE residuals, and sparsity.
In the above loss, Eq. \eqref{eq:L_data} contains all the data fitting terms for the \textit{data variables}. 
The outputs for the \textit{data variables} along with the outputs of the \textit{equation variables} are used to define the loss on the learned vector fields of the \textit{equation variables} in Eq. \eqref{eq:L_phy}.
Eq. \eqref{eq:L_l0} induces sparsity in MUSIC by considering the $\ell_0$ norm of all the model parameters (weights and biases).
The pseudocode for the proposed model and algorithm is given in Algorithm \ref{alg:Loss}.
In order to enforce sparsity in the, we consider both weight and neuron sparsity, explained briefly below.

\textbf{Weight (Unstructured) sparsity:} We use a thresholding approach as implemented in \citep{blumensath2009iterative,foucart2011hard} where the model weights are pruned after every (few) iterations. 
To train MUSIC, we first optimize the dense network for a few epochs (about 10k epochs in our result(s)). 
This is done to ensure that the model does not collapse or lose important weights right from the beginning.
Then every validation step, a hard thresholding weight pruning approach is adapted to zero out the parameters below a certain threshold level (or only the top $k$ weights, by magnitude are kept and the rest are zeroed out). 
In this case, the loss function only contains Eq. \eqref{eq:L_data} and \eqref{eq:L_phy}, and Eq. \eqref{eq:L_l0} is induced through iterative thresholding.

\textbf{Neuron (Structured) sparsity:} We modify Eq. \eqref{eq:L_l0} in the loss function directly so that gradient based optimizers can be used in model training. 
Since the $\|\mathbf{\Theta}\|_0$ is not differentiable we consider a continuous approximation of $\|\mathbf{\Theta}\|_0$.
In particular, we use a hard-concrete distribution approximation of $\|\mathbf{\Theta}\|_0$ as proposed in \citep{louizos2017learning} which is briefly summarized here.
For input-output pairs $\{(x_1, y_1), \dots, (x_N, y_N)\}$, a neural network $h(x_i; \Theta)$ with network parameters $\Theta$, the loss function can be defined as
\begin{equation}
R(\Theta) = \frac{1}{N} \sum_{i=1}^{N} \|h(x_i; \Theta)- y_i\|_2^2 + \lambda \|\Theta\|_0, 
\quad
\|\Theta\|_0 = \sum_{j=1}^{|\Theta|} \mathbb{I}[\Theta_j \neq 0]\nonumber.
\end{equation}
Since $\|\Theta\|_0$ is not differentiable, the goal is to approximate it with a continuous term so that gradient based optimizers can be directly applied to the loss function. Suppose
\[
\Theta_j = \tilde{\Theta}_j z_j, \quad z_j \in \{0,1\}, \quad \tilde{\Theta}_j \neq 0, \quad 
\|\Theta\|_0 = \sum_{j=1}^{|\Theta|} z_j,
\]

where $z_j$ are binary “gates” denoting if a parameter is present. 
\[
\text{If } q(z_j \mid \pi_j) = \text{Bern}(\pi_j)
\]
\begin{equation}
R(\tilde{\Theta}, \pi) = \mathbb{E}_{q(z \mid \pi)} \Bigg[
\frac{1}{N} \sum_{i=1}^{N} \mathcal{L}\big(h(x_i; \tilde{\Theta} \odot z), y_i\big) 
+ \lambda \sum_{j=1}^{|\Theta|} \pi_j \Bigg].\nonumber
\end{equation}

\begin{equation}
(\tilde{\Theta}^*, \pi^*) = \arg \min_{\tilde{\Theta}, \pi} R(\tilde{\Theta}, \pi)\nonumber
\end{equation}
where Bern($\cdot$) denotes the discrete Bernoulli distribution with parameter $\pi$ and $\odot$ denotes Hadamard product.
Note that since discrete Bernoulli gates are not differentiable, the \textbf{hard concrete distribution} along with some reparameterizations is used to obtain
\[
z_j = \min(1, \max(0, s_j)), \quad s_j = \sigma\Bigg( \frac{\log u_j - \log(1-u_j) + \log \alpha_j}{\beta} \Bigg) \cdot (r-l) + l,
\]
where $u_j \sim \mathrm{Uniform}(0,1)$, $\sigma(\cdot)$ is the sigmoid function, $\alpha_j$ is a trainable parameter that controls the probability of the gate being “on” and $\beta, r, l$ denote the temperature, right and left stretch interval respectively that control the interval and strength of the number of neurons that should be kept `` on".
Note that the weights do not actually become `zero' since the model needs them to pass the gradients. The gates are simply turned ``off" which is the same as the neurons being treated as `zero'. 
However, once model training is complete, the ``off" neurons can be made zero using model pruning. 

In this paper, we particularly focus on coupled systems i.e., when $n=2$. Other higher order cases can be formulated similarly. 
After training, we compute the relative $\ell_2$ loss for both the solutions variables, as well as the full-field (stacked) relative $\ell_2$ loss. 
For the true solutions $u_1$, $u_2$ and their learned solutions $\widehat{u}_1, \widehat{u}_2$ the relative errors are calculated as,
\begin{equation}
\begin{aligned}
    &\text{Error}_{(u_1,\widehat{u}_1)} = \dfrac{\|u_1-\widehat{u}_1\|_2}{\|u_1\|_2} \,\,\text{and}\,\,\text{Error}_{(u_2,\widehat{u}_2)} = \dfrac{\|u_2-\widehat{u}_2\|_2}{\|u_2\|_2},\\
    &\text{Error}_{([u_1,u_2],[\widehat{u}_1\widehat{u}_2])} = \dfrac{\|\mathbf{z}-\widehat{\mathbf{z}}\|_2}{\|\mathbf{z}\|_2}\,\,\,\text{where}\,\,\,\mathbf{z} = [u_1,u_2]^T\,\,\,\text{and}\,\,\,\mathbf{\widehat{z}} = [\widehat{u}_1,\widehat{u}_2]^T.
    \end{aligned}
\end{equation}
\begin{algorithm}
\caption{Learning Sparse Multitask Models to Solve Coupled PDEs with Disjoint Priors}
\begin{algorithmic}[1]  % The number in the brackets is the line numbering
    \STATE Initialize: \texttt{Net, hid\_dim, layers, max\_epoch, weight sparsity k, Method, $\lambda_1$, $\lambda_2$, $\lambda_0$, z$\in\{0,1\}$, Sigmoid function $\sigma$, trainable neuron $\ell_0$ parameter $\alpha$.}
    % \pause
    \FOR{\texttt{epoch} $<$ \texttt{max\_epoch}}
    % \pause
    \IF{Method $==$ `weight-L0'}   
    \STATE $\widehat{u},\widehat{v} = $\texttt{Net}($\mathbf{x},t,$,\texttt{hidden\_dim, layers})
    \STATE PDE$_{\widehat{v}}$ = $F(\widehat{u},\widehat{v})$
    \STATE Define \texttt{Loss} = $\lambda_1\|u-\widehat{u}\|_2^2 + \lambda_2\|\widehat{v}_t - F(\widehat{u},\widehat{v})\|_2^2$ 
    \STATE Optimize
        \IF{\texttt{epoch}$\%$\texttt{val\_step}$==0$}
        % \FOR{\texttt{params} in \texttt{Net}}
            \STATE Keep \texttt{params\_nnz} = Top$_k$(\texttt{params}) every few \texttt{epochs}.
            \ENDIF
% \pause
        \ELSE
        \STATE $\widehat{u},\widehat{v} = $\texttt{Net}($\mathbf{x},t, z$\texttt{hidden\_dim, layers}) \STATE PDE$_{\widehat{v}}$ = $F(\widehat{u},\widehat{v})$
          \STATE  Define Loss = $\lambda_1\|u-\widehat{u}\|_2^2 + \lambda_2\|\widehat{v}_t - F(\widehat{u},\widehat{v})\|_2^2$ + $\lambda_0\sigma(\alpha)$
          \STATE Optimize
        \ENDIF
    \ENDFOR
    \STATE \textbf{Return} Trained \texttt{Net}
\end{algorithmic}\label{alg:Loss}
\end{algorithm}

\section{Learning Solutions from Low Dimensional Priors for Coupled 2D Systems}
\subsection{1D Shallow Water Equations (SWE)}\label{sec:swe}
In this section, we discuss the model formulation and results for learning the full-field solution of the shallow water equations (SWE). The SWE is a system of coupled PDEs governing fluid
flow in the oceans, coastal regions, estuaries, rivers, channels, etc. Derived from the Navier-Stokes equations, which
describe the motion of fluids, a key characteristic of the SWE system is that the horizontal length scale is much greater than the vertical depth. The assumption here is that vertical velocity is negligible, which is reasonable since we are consider the flow of shallow water. 
Given $x\in \Omega$, where $\Omega\subset\mathbb{R}$ denotes the (spatial) input domain, suppose $h$ denotes the height of water, and $u$ denotes the horizontal velocity. For $g = 9.81$ $ms^{-2}$ (gravitational constant) and bed surface $z_b$, the (rearranged) 1D SWE is given by
\begin{equation}\label{eq:swe}
\begin{aligned}
    h_t + (hu)_x &=0\\
    (hu)_t + \left(hu^2 + \frac{1}{2}gh^2\right)_x + g(z_b)_x&=0 .\\
 \end{aligned}
 \end{equation}
We apply this system to the well-known dam break problem \citep{crowhurst2013numerical}. For $\Omega\times [0,1]$, where $\Omega = [0,10]$, assuming bed surface $z_b = 0$, we obtain the full-field solution of this system using the initial conditions, 
\begin{equation}
\begin{aligned}
     h(x,0) & =  \begin{cases} 
      1, & \text{if } x\in [0,5] \\
      0, & x\in (5,1]
   \end{cases}\\
   u(x,0) & = 0 \,\,\forall \, x\in [0,10].
   \end{aligned}
\end{equation}
The boundary conditions are chosen such that there is no flow in or out at the boundaries i.e., a zero-gradient (Neumann) boundary conditions with $\dfrac{\partial h}{\partial x} = \dfrac{\partial u}{\partial x} = 0$ for $x\in\partial\Omega$, where $\partial\Omega$ denotes the boundary. The Local Lax-Friedrichs (Rusanov flux) finite volume method is used to obtain the solutions with $\Delta x = 0.5$ and $\Delta t = 0.001$. The true solutions at $t=\{0, 0.2, 0.4, 0.6, 0.8, 1.0\}$ are given below in Figure \ref{fig:swe_true}.

\begin{figure}[h!]
    \centering
    \includegraphics[scale = 0.25]{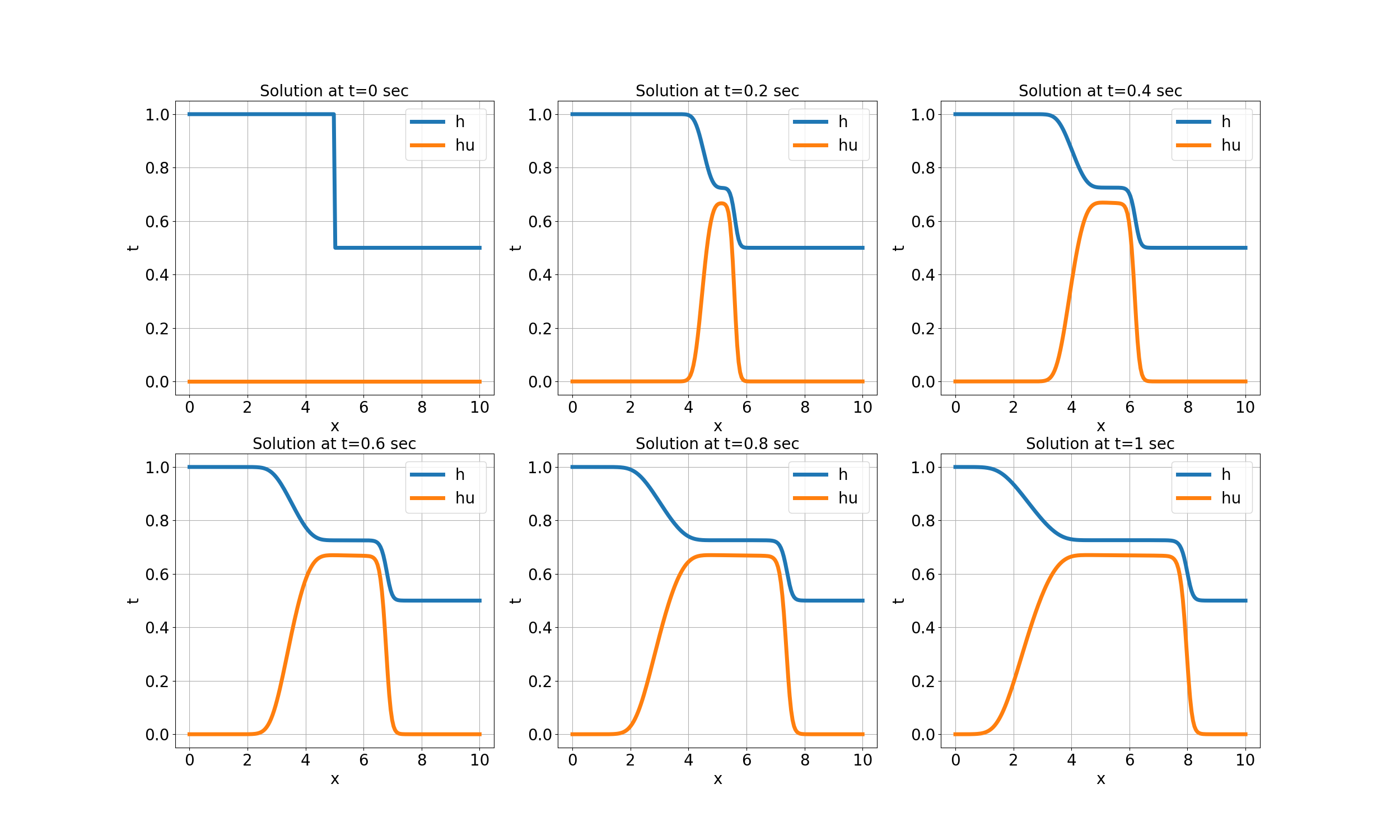}
    \caption{True solutions $\{h,hu\}$ of the dam break problem modeled by SWE at $t=\{0,0.2,0.4,0.6,0.8,1.0\}$.}
    \label{fig:swe_true}
\end{figure}

% \subsubsection{Learning solutions $[h,hu]^T$ given equation for $h$ and data for $hu$}
Our goal is to use mutually disjoint priors from SWE system to learn the full-field solution. The variable represented by a less complex vector field is chosen as the equation variable. 
For complex vector fields, the solution variable is treated as a data variable. 
Note that the proposed method is also applicable to cases when the data and equation variable are switched i.e., the complex equation is used for physics regularization and solution to the easier equation is used for data fitting (see Section \ref{sec:switched-data-eq-var}).
Clearly, the equation for $h$ is less complex and is thus used for physics regularization (equation variable) and $hu$ is the data variable.
% Note that if data for $u$ is available, then it can also directly be for training instead of $hu$.
Let $N_s$ and $N_t$ denote the number of training points on the spatial and temporal domain, respectively and $N_{ic}$ denote the number of spatial data available at $t=0$. Suppose $\widehat{h}_{ik}$ and $(\widehat{hu})_{ik}$ denote the learned solutions at $(x_i,t_k)$, then the loss function for optimizing the weights with structured $\ell_0$ sparsity is given by 
\begin{equation}
\begin{aligned}
    \text{Loss} &= \dfrac{1}{N_sN_t}\sum\limits_{i=1}^{N_s}\sum\limits_{k=1}^{N_t}
    \left(\left\|(hu)_{ik} - (\widehat{hu})_{ik}\right\|_2^2 +  \left\| (\widehat{h}_t)_{ik} + ((\widehat{hu})_t)_{ik}\right\|_2^2\right)\\ & + \dfrac{1}{N_{ic}}\sum\limits_{i=1}^{N_{ic}}\left\|h_i - \widehat{h}_i\right\|_2^2 + \lambda_0\sum\limits_{j=1}^{|\text{Neurons}|}\sigma(\alpha_j),
    \end{aligned}
\end{equation}
where $\sigma$ denotes the (logistic) sigmoid function and $\alpha_j$s are the trainable parameter controlling the probability of the neurons being ``on" (see Section \ref{sec:model-framework} for details). 
We use one fully connected neural network with 2D vector output to represent the learned $[h,hu]^T$ i.e., the model utilizes the shared weights in the hidden layers to learn the full-field solution.
A structured/neuron sparsity is used by considering a continuous approximation of the $\ell_0$ norm using the hard-concrete distribution.
We normalize the data (and hence modified SWE based on normalized variables) where all the input and output variables are constrained within [0,1] using the min-max normalization. 
A small subset of $hu$ are randomly sampled from the spatio-temporal domain to construct the training dataset. 
We use 80\% of the data for training and 20\% for validation and hyperparameter tuning. 
The ReLU activation function is used in the model majorly for two reasons. One, the solution is always positive and second, if we look at the initial condition of $h$, we clearly see it to be a step function.
We use the ADAM optimizer with learning rates tuned between $10^{-2}$ and $10^{-4}$ over $10^4$ iterations.
The $\ell_0$ regularization parameter $\lambda_0$ is tuned between $10^{-4}$ and $10^{-10}$ using the validation data.
% In order to check for model robustness, we train the model with 3-5 random seeds (initializations) and report the mean and standard deviations of the full-field average relative $\ell_2$ errors.

\begin{table}[h!]
    \centering
    \begin{tabular}{|c|c|c|c|}
    \hline
      \backslashbox{Neurons}{Layers} & 2 & 4 & 6 \\
      \hline
       20  & 5.86\% $\pm$ 2.87\% & 3.96\% $\pm$ 1.12\% &
4.41\% $\pm$ 1.44\%\\
Sparsity (\%) &12.64\% & 7.93\% & 6.81\%\\
% Training Time (s) & 82&107 &133\\
       \hline
       50  & 3.71\% $\pm$ 2.51\% & 4.22\% $\pm$ 3.10\% & 6.59\% $\pm$ 3.28\% \\
      Sparsity (\%) & 5.57\% & 3.26\% & 2.77\% \\
      % Training Time (s) & 82&108 &136\\
       \hline
    \end{tabular}
    \caption{Full-field average relative $\ell_2$ error (in percentage) with $\ell_0$ neuron sparsity for SWE system. The model was trained using $N_s = 100$, $N_t = 800$.}
    \label{tab:swe_neuron_vs_layers}
\end{table}

In Table \ref{tab:swe_neuron_vs_layers} we report the average relative $\ell_2$ error (in percentage) for learning the full-field solutions $[h,hu]^T$. 
The size of sampled dataset, for training and validation is $N_s*N_t = 80,000$ ($N_s=100$, $N_t=800$) of which 80\% is used for training and the rest 20\%is used as a validation set to tune the learning rate. 
Thus, the training dataset consists of about 64000 spatio-temporal points (about 32\% of the full dataset which consists of 200 spatial points and 1000 temporal points) used as collocation points (physics-based regularization) as well as data fitting.
We use different levels of model complexity in terms of the number of hidden layers and the hidden dimension (number of neurons per layer). 
We see that as the model model complexity increases, generally the error goes down, however plateauing after a particular threshold. For example, when use 4 layers, the relative error of \textit{most} models remain below 5\% irrespective of the number of hidden nodes. 
\begin{figure}[h!]
    \centering
    \includegraphics[width=1.0\linewidth]{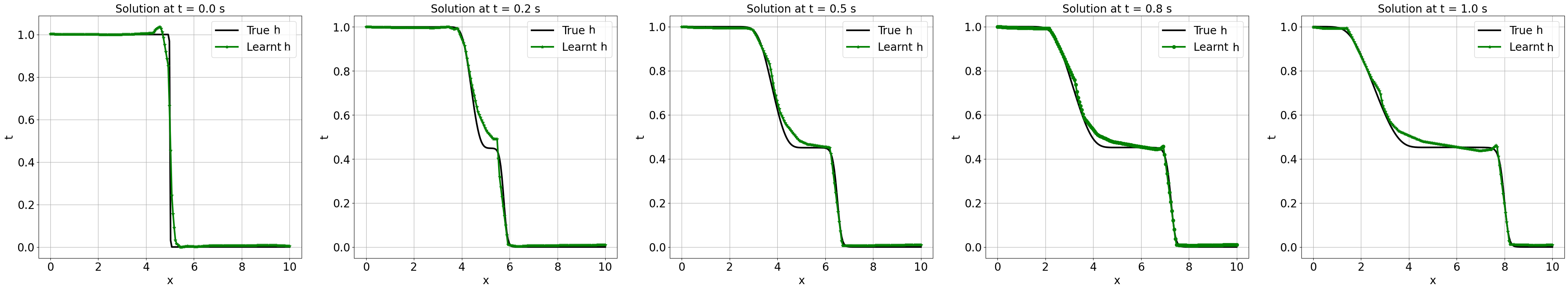}\\
    \includegraphics[width=1.0\linewidth]{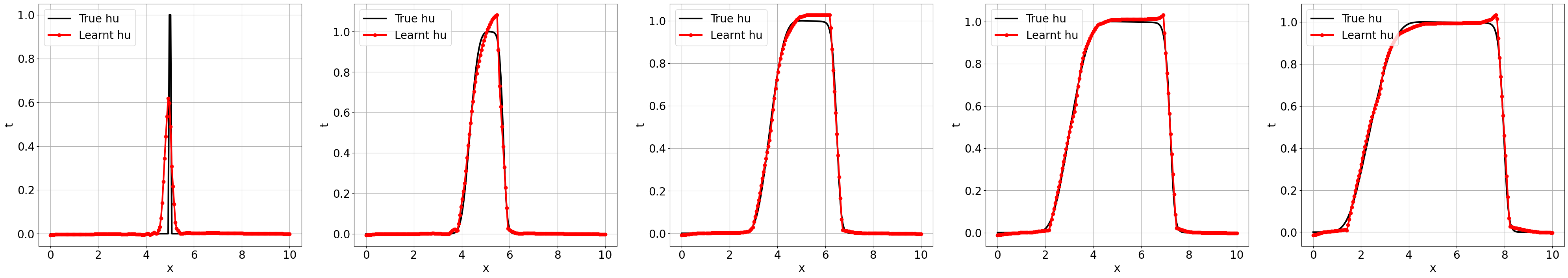}
    \caption{True and learned solutions $h$ and $hu$ using a 4 layer model with 20 neurons with $N_s=100$ and $N_t=800$. \textbf{Top row}: True and learned solution for equation variable $h(x,t)$. \textbf{Bottom row}: True and learned solution for data variable $(hu)(x,t)$.}
    \label{fig:swe-sol-best}
\end{figure}
However, if the model complexity increases too much, the error increases as seen in the case of a 6 layer model with 50 hidden dimension.
In this case, the model is so overparameterized in comparison to number of training data that even with appropriate $\ell_0$ regularization it ends up overfitting.
Thus, it is important to use a model structure based on the amount of input data available.
The plot of the learned and true solutions for a 4 layer model with 20 hidden dimension is given in Figure \ref{fig:swe-sol-best}.

\subsubsection{Effect of Noise on Model Performance}
\begin{table}[h!]
    \centering
    \begin{tabular}{|c|c|c|c|}
    \hline
      \backslashbox{$N_s$}{Noise} & 2\% & 5\% & 10\%\\
     \hline
       50  & 4.80\% $\pm$ 2.30 \% & 4.77\% $\pm$ 1.67 \%& 5.29\% $\pm$ 0.57 \%\\
       \hline
       100  & 4.83\% $\pm$ 1.87\% & 5.07\% $\pm$ 2.55\% & 5.62\% $\pm$ 0.81\%  \\
       \hline
       150  & 4.27\% $\pm$ 1.05\%& 4.50\% $\pm$ 0.21\% & 6.27\% $\pm$ 0.80\%\\
       \hline
    \end{tabular}
    \caption{Full-field average relative $\ell_2$ error (in percentage) for SWE system with different levels of noise and $N_{s}$. A 4 layer fully connected neural network model with 20 neurons was trained using $N_t = 800$. The sparsity for the model was at 7.93\%.}
    \label{tab:swe_noise_vs_data}
\end{table}
In order to check the robustness of MUSIC with respect to different levels of noise, we corrupt the data variable $hu$ with different levels of noise. 
Noise levels do not affect $h$ since it learns directly from its governing equation.
The average relative $\ell_2$ error of the solutions for different levels of noise versus the number of spatial points used for training is given in Table \ref{tab:swe_noise_vs_data}. 
The model is trained by fixing the number of temporal points to 800, the total number of hidden layers to 4 and a width of 20.
Similar to previous cases, a 80-20 split for training and validation was employed to tune the hyperparameters.
As expected, when the noise levels are low, the model performs well even with merely 50 spatial points, improving when we increase the number of spatial data to 150. However, as the noise level increases, the model performance starts to deteriorate. Further, for higher noise levels at 10\%, an increase in the number of spatial data may not necessarily improve the performance as more training data also introduces more noise into the model, which could lead to overfitting with the noise.

% of low accuracy in the first two rows of Table \ref{tab:predict_swe_future} could also be due to the prediction interval being much larger than that of the last row, leading to larger errors for farther time points. 

% \subsubsection{Identification of System using SINDy from Parameterized Full-Field Solution}
\subsection{Fitzhugh-Nagumo System}
\begin{figure}[h!]
    \centering
    \includegraphics[scale=0.35]{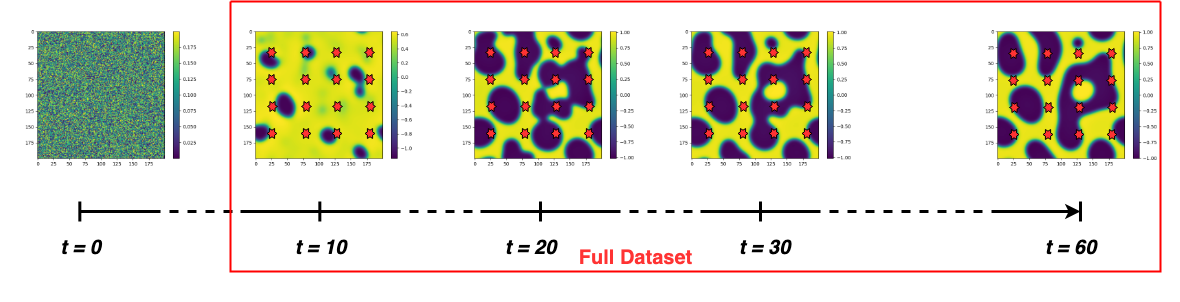}
    \caption{Dataset creation for the FN system. The red dots illustrate the evenly spaced points chosen in the spatial domain. The full dataset is built by taking 50 evenly spaced temporal solutions i.e., solutions at $t=\{10,11,12,\dots,59,60\}$.}
    \label{fig:FNdataset}
\end{figure}

In this section, we consider the Fitzhugh-Nagumo (FN) type reaction-diffusion system, in a 2D spatial domain $\Omega\subset \mathbb{R}^2$ with periodic boundary conditions, whose governing equations are two coupled PDEs described as

    \begin{align}
u_t &= \gamma_u \Delta u + u - u^3 - v + \alpha, \notag \\
v_t &= \gamma_v \Delta v + \beta(u - v), \label{eq:fhn}
\end{align}
where $u$ and $v$ represent two interactive components, $\gamma_u$ and $\gamma_v$ are diffusion coefficients, $\alpha$ and $\beta$ are the reaction coefficients, and $\Delta$ is the Laplacian operator \citep{fitzhugh1961impulses,chen2021physics,nagumo2007active}. 
The FN equations are generally used to describe pattern formation in biological neuron activities excited by an external stimulus $\alpha$. The system exhibits an activator-inhibitor system where one equation boosts the production of both components while the other equation dissipates their new growth \citep{chen2021physics}.
We generate the ground truth data by solving the system using a finite difference method with initial conditions taken as random vectors such that $u(x,y,0),v(x,y,0) \sim \mathcal{N}(\mathbf{0},0.2\mathbf{I}_2)$, where $\mathbf{0}$ and $\mathbf{I}_2$ denote the zero vector and $2\times 2$ identity matrix respectively.
The system is solved using $dx = dy = 0.5$ for $x,y \in [0,100]$ and $dt = 0.0005$ for $t\in [0,60]$. We select 100 evenly spaced data in time for $t\in [10,60]$ and 50  evenly spaced spatial data in both $x$ and $y$ direction to create the full-field ground truth dataset of size (100,100,50). 
Full dataset building from the spatiotemporal domain is described in Figure \ref{fig:FNdataset}.
We discard the solutions for $t\in [0,10)$ as the solutions are mostly (full of) noise and patterns are yet to form.
Set $u$ as the data variable and $v$ as the equation variable. Let $N_{s_x}$ and $N_{s_y}$ denote the training points in the 2D spatial domain, $N_t$ denote the number of temporal training points, and $N_{ic}$ denote the total number of spatial data at $t=0$. For each $i=1,\dots, N_{s_x}$, $j=1,\dots, N_{s_y}$ and $k = 1,\dots,N_t$, suppose $\widehat{u}_{ijk}$ and $(\widehat{v})_{ijk}$ denote the learned solutions at $(x_i,y_j,t_k)$, then the loss is given by 
\begin{align}
    \text{Loss} = &\dfrac{1}{N_{s_x}N_{s_y}N_t}\sum\limits_{i=1}^{N_{s_x}}\sum\limits_{j=1}^{N_{s_y}}\sum\limits_{k=1}^{N_t}
    \left(\left\|u_{ijk} - \widehat{u}_{ijk}\right\|_2^2 +  \left\| (\widehat{v}_t)_{ijk} - \gamma_v\Delta\widehat{v}_{ijk} - \beta (\widehat{u}_{ijk} - \widehat{v}_{ijk})\right\|_2^2\right) \\
    &+ \dfrac{1}{N_{ic}}\sum\limits_{i=1}^{N_{ic}}\left\|v_i - \widehat{v}_i\right\|_2^2 + \lambda_0\sum\limits_{j=1}^{|\text{Neurons}|}\sigma(\alpha_j),\nonumber
\end{align}
where $\sigma$ denotes the (logistic) sigmoid function and $\alpha_j$s are the trainable parameter controlling the probability of the neurons being ``on" (see Section \ref{sec:model-framework} for details).
A fully connected multitask neural network with shared weights (similar to previous SWE system) is used to learn the full-field solution.
Similar to the previous case, we first normalize the input and output data using min-max normalization.
A 80-20 training-validation split is used to train the model using ADAM optimizer and tune the learning rate between $10^{-2}$ and $10^{-4}$. The number of collocation points used were the same as the number of training points.
Note, this technique of using the same number of collocation points as the data is different from previous PIML approaches \citep{chen2021physics,raissi2019physics}.
We enforce a $\ell_0$ sparsity on each layer to activate neuron sparsity. The regularization parameter $\lambda_0$ is tuned from the validation set between $10^{-2}$ and $10^{-8}$.
% \subsubsection{Case I: Random Initialization}

\begin{table}[h!]
    \centering
    \begin{tabular}{|c|c|c|}
    \hline
        \backslashbox{Neurons}{Layers} & 4 & 6 \\
        \hline
        100 & 11.51 \% $\pm$ 5.32 \% & 9.07 \% $\pm$ 1.23\% \\
        Sparsity (\%) & 1.65\% & 0.4\%\\
        % Time taken (s) &2104 & 6670\\
        \hline
        200 & 7.97 \% $\pm$ 0.79\% & 7.66\% $\pm$ 1.43\% \\
        Sparsity (\%) & 0.77\% & 0.2\%\\
        % Time taken (s) &4576 & 15000\\
        \hline
    \end{tabular}
    \caption{Full-field average relative $\ell_2$ error (in percentage) for FN system. The model was trained using $N_{s_x}*N_{s_y} = 5000$, $N_t = 40$.}
    \label{tab:fhn_neurons_layers}
\end{table}

\begin{figure}[h!]
     \centering
     \begin{subfigure}[b]{1.0\textwidth}
         \centering
         \includegraphics[width=\textwidth]{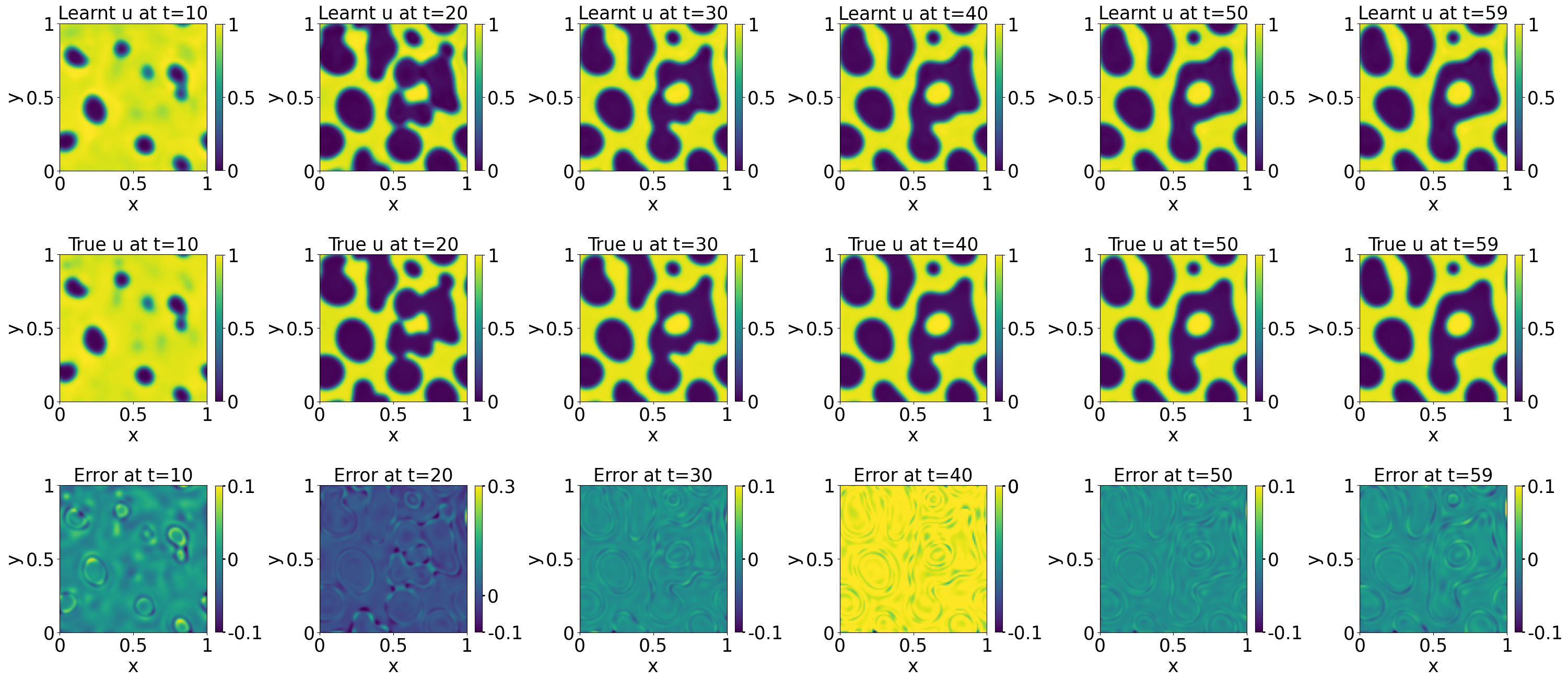}
         \caption{Predicted solutions (top row), true solution (middle row) and error (bottom row) for $u$.}
         \label{fig:fhn1_u}
     \end{subfigure}
     % \vspace{-5mm}
     % \hfill
     \begin{subfigure}[b]{1.0\textwidth}
         \centering
         \includegraphics[width=\textwidth]{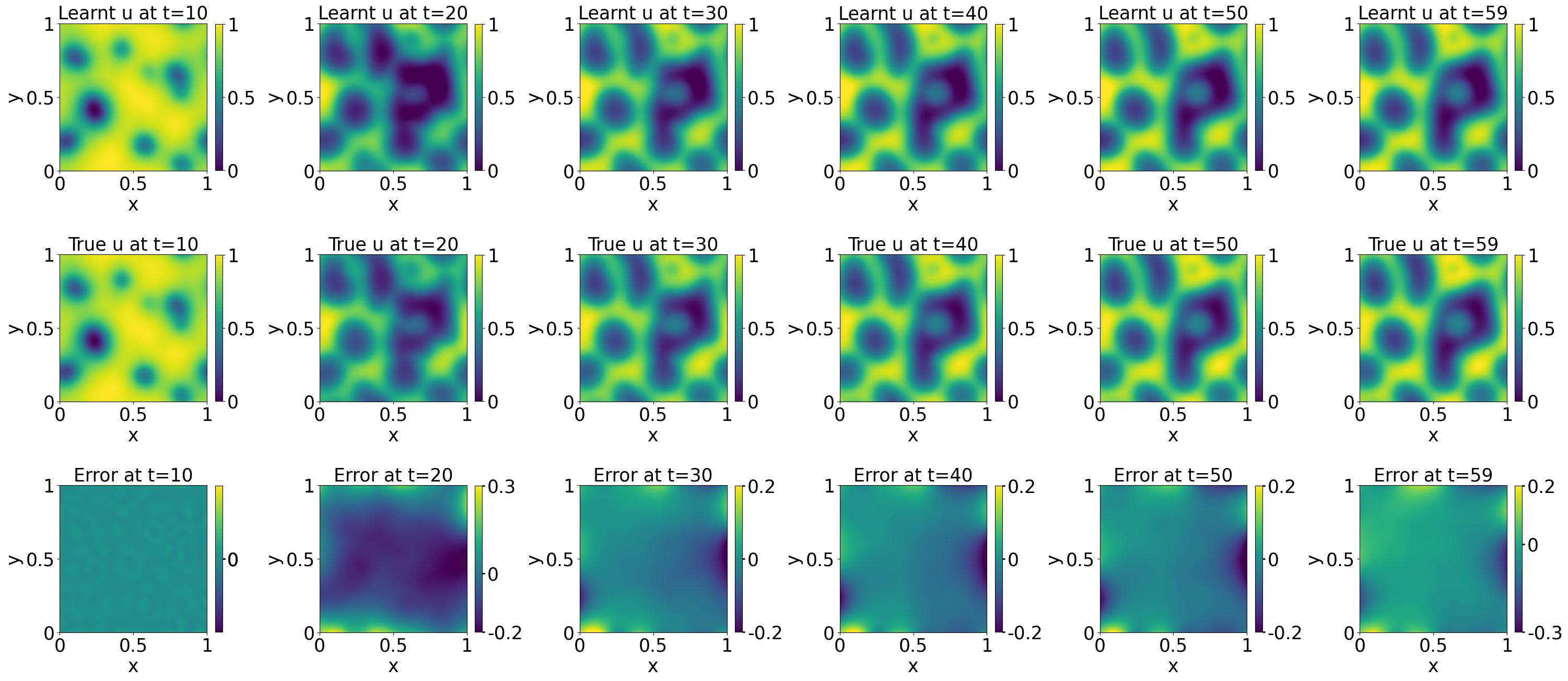}
         \caption{Predicted solutions (top row), true solution (middle row) and error (bottom row) for $v$.}
         \label{fig:fhn1_v}
     \end{subfigure}
        \caption{Solution to the FH system with random initialization learned using a 4 layer fully connected neural networks using 200 neurons with $N_{s_x}*N_{s_y} = 5000$ and $N_t = 40$. Note the scales for the error are different for each plot.}
        \label{fig:fhn1_uv}
\end{figure}

The performance of the model with respect to different layers and hidden dimension is given in Table \ref{tab:fhn_neurons_layers}. 
The sparsity level (number of nonzero weights in the trained model) is also mentioned for each model.
Additionally, for the best performing model, which is the one with 6 layers and 200 neurons (considering both error and its standard deviation) we plot the full-field learned solutions $u$ and $v$ at various timesteps in Figure \ref{fig:fhn1_uv}.
Each plot in Figure \ref{fig:fhn1_u} and Figure \ref{fig:fhn1_v} depicts the learned solution (top row), the true solution (middle row) and the difference between true and learned solution (bottom row).
For most of the cases the error is around $\pm$ 0.1, with the worse case scenario having an error of $\pm$ 0.3 either at the spatial boundary of along the contours of the patterns.
This happens as we do not explicitly enforce the boundary conditions during training.
As for the errors around the pattern contours, this happens as the machine learning model learns a continuous function to represent the patterns leading to a higher error.
At $t=10$ (initial condition corresponding to $t=0$ during model training), since we use initial conditions and the physics of the equation variable $v$, the model perfectly learns the initial solution while for the data variable $u$ the error is between $\pm$ 0.1.
The error plots show that for most all the cases, $\widehat{u}-u$ and $\widehat{v}-v$ remains very close to zero. 
We also plot the average error evolution across each timestep in Figure \ref{fig:fhn-error-evolve-clean} for the 6 layers-100 hidden dimension model. Upon examining the relative errors at each timestep, we found that for data variable $u$, since data is used across the time domain, the errors remain fairly constant except at a few points around $t=20$, which when tallied with the true plots can be attributed to the time points around which some disconnected patterns start joining. 
However, eventually the errors plateau out.
For equation variable $v$, due to patterns in the earlier timesteps, the errors are relatively higher. However, as the patterns start to form and become more well-defined, the model error falls, plateauing around/ below $10\%$. 

\subsubsection{Effect of Noise and Timesteps in Model Training}
\begin{figure}[h!]
     \centering
     \begin{subfigure}[b]{0.3\textwidth}
         \centering
         \includegraphics[width=\textwidth]{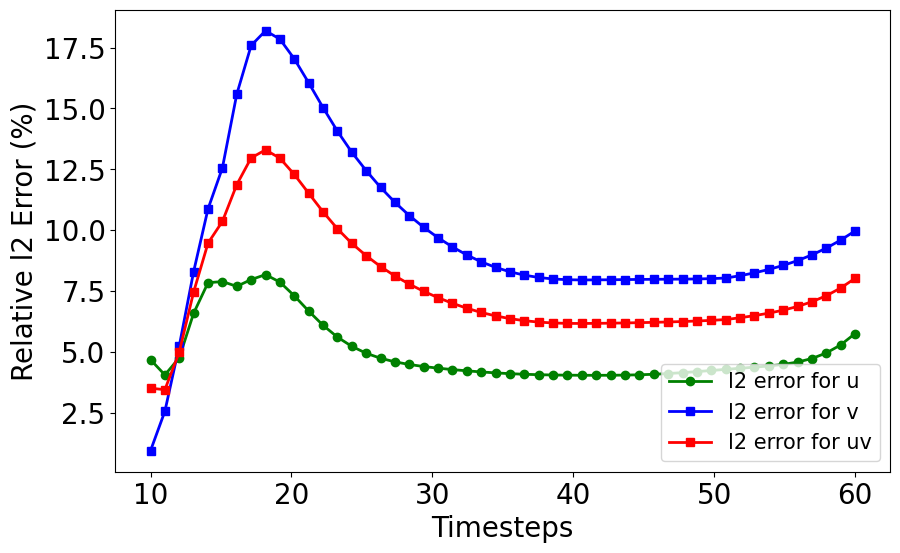}
         \caption{Noise level: 0\%}
         \label{fig:fhn-error-evolve-clean}
     \end{subfigure}
     \begin{subfigure}[b]{0.3\textwidth}
         \centering
         \includegraphics[width=\textwidth]{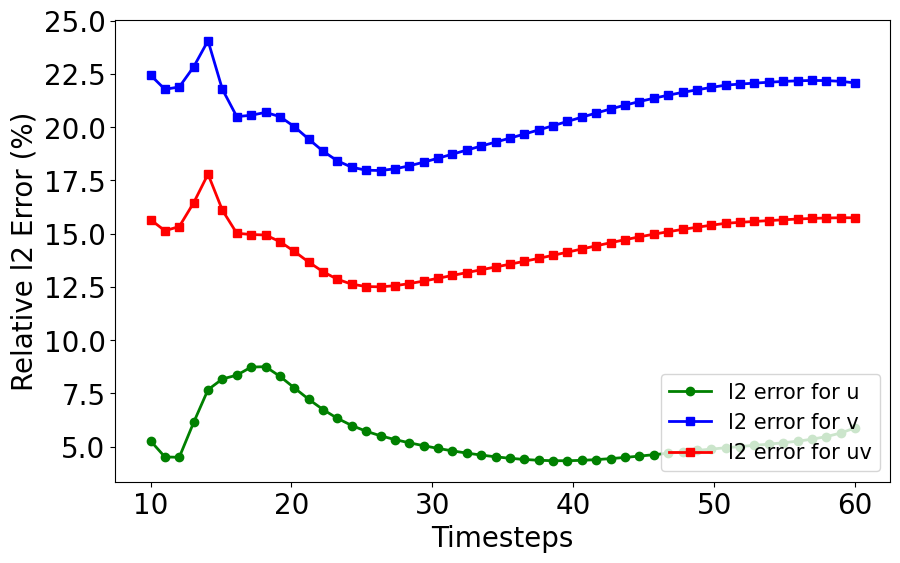}
         \caption{Noise level: 2\%.}
         \label{fig:fhn-error-evolve-0.05}
     \end{subfigure}
          \begin{subfigure}[b]{0.3\textwidth}
         \centering
         \includegraphics[width=\textwidth]{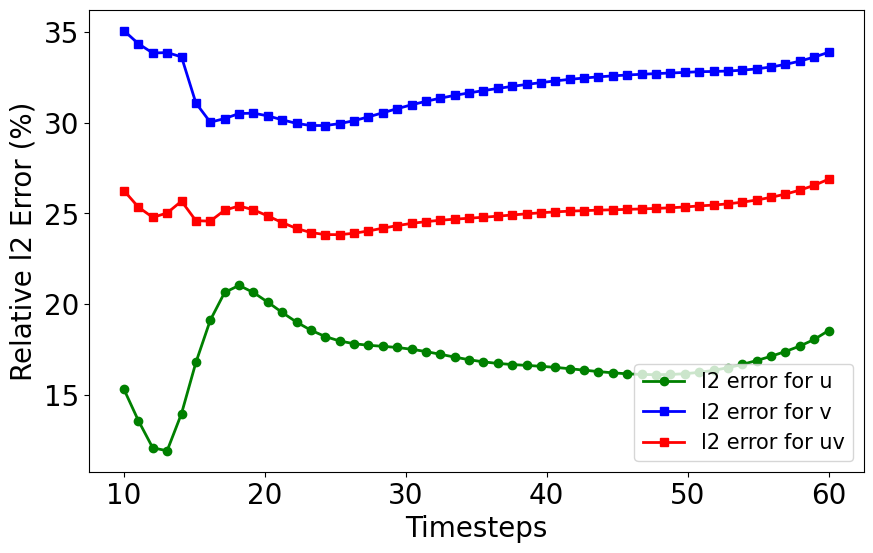}
         \caption{Noise level: 10\%.}
         \label{fig:fhn-error-evolve-0.1}
     \end{subfigure}
        \caption{Error evolution of learning solutions to the FN system across different timesteps and noise levels. $N_{s_x}*N_{s_y} = 5000$ and $N_t = 40$.}
        \label{fig:fhn-error-evolve}
\end{figure}

To test the model's robustness against noise, we train the model with different numbers of spatial points with different noise levels. The full-field average relative $\ell_2$ error is given in Table \ref{tab:fhn1-noise-error}. As expected, we see a degradation in performance when noise level increases. 
Upon analyzing the errors made across each timestep (see Figure \ref{fig:fhn-error-evolve}), we found an interesting pattern of errors for different noise levels. 
When the error is low at 2\%, the error evolution over time shows a similar pattern as that of clean data.
The error made in learning is the solution at initial timestep $t=10$ is similar to the overall average error, followed by an increase and plateauing of errors after a certain number of timesteps.
However, when the noise increases to 10\%, the initial solution at $t=10$ becomes much closer to pure noise (as the entire dataset was initially simulated using random initial conditions) leading to highest error for learning equation variable $v$ at $t=10$. 
For the data variable $u$, the error evolution pattaern remains fairly similar, with an overall increased errors based on noise level.
However, as the patterns become more structured, the errors remain more or less the same thereafter.
\begin{table}[h!]
    \centering
    \begin{tabular}{|c|c|c|c|}
    \hline
      \backslashbox{$N_{s_x}*N_{s_y}$}{Noise} & 2\% & 5\% & 10\%\\
     \hline
       500  & 16.28\% $\pm$ 2.73 \% & 21.88\% $\pm$ 1.28 \%& 27.98\% $\pm$ 0.68 \%\\
       \hline
       1000  & 14.98\% $\pm$ 0.79\% & 22.39\% $\pm$ 1.36\% & 24.87\% $\pm$ 0.21\%  \\
       \hline
    \end{tabular}
    \caption{Full-field average relative $\ell_2$ error (in percentage) for FN system with different levels of noise and $N_{s_x}*N_{s_y}$. A 6 layer fully connected neural network model with 100 neurons was trained using $N_t = 40$. The sparsity for the model was at 0.4\%.}
    \label{tab:fhn1-noise-error}
\end{table}
Also, depending on the level of noise, adding more spatial points may help to improve the accuracy as depicted in Table \ref{tab:fhn1-noise-error} where generally the model with $N_{s_x}*N_{s_y} = 1000$ performs better than using $N_{s_x}*N_{s_y} = 500$.
Although it may also be interesting to note that for noisy data, a higher number of spatial points for training may not always be beneficial since it also introduces more noise in the training process.

\begin{figure}[h!]
     \centering
     % \begin{subfigure}[b]{0.3\textwidth}
         % \centering
         \includegraphics[width=0.6\textwidth]{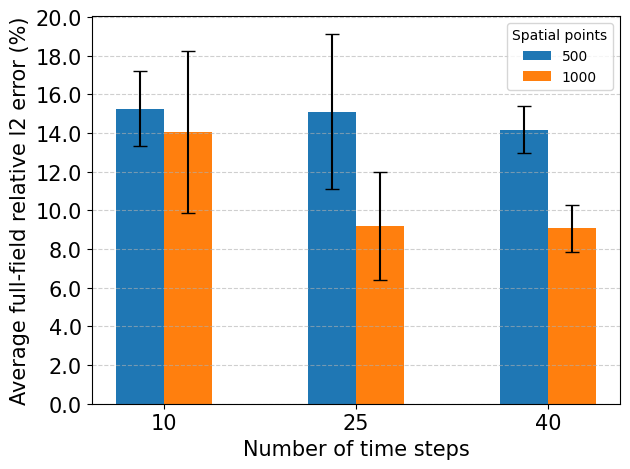}
         \caption{Full-field averae $\ell_2$ error (in percentage) for learning solutions to the FN system with 6 layers, 100 neurons and varying $N_t$ and $N_s$.}
         \label{fig:Nt-vary-fhnL0}
\end{figure}

For all the simulations above, we used $40$ randomly chosen timesteps from $\{1,\dots,50\}$ to train the model.
We also tested how the model would perform if we reduced the number of timesteps used in training. 
For testing the affect of $N_t$, we trained a model with 6 layers, 100 neurons and varied $N_t = \{10,25,40\}$ and $N_s = \{500,1000\}$. 
A histogram of the full-field $\ell_2$ errors made for each $N_t$ and $N_s$ is given in Figure \ref{fig:Nt-vary-fhnL0}.
When very less number of spatial points are used i.e., $N_{s_x}*N_{s_y} = 500$, the error remains over 14\% irrespective of the number of timesteps used.
However, if we increase the number of spatial points to 1000, then we clearly see an improvement from using $N_t=10$ to $N_t=25$ with an improvement of almost 6\%. 
It is also interesting to note that the errors for both $N_t=25$ and $N_t=40$, are very similar indicating that as less as $N_t=25$ timesteps may be sufficient for training the model to obtain an accuracy of below 10\%.
Figure \ref{fig:Nt-vary-fhnL0} highlights the importance of having a larger amount of spatial data compared to temporal data. It shows that increasing the number of spatial data points can significantly reduce the error, even when the number of temporal data points is relatively small.

% \subsubsection{Case II: Initial condition from \citep{chen2021physics}}
% In this section, we use the system of equations described in Eq. \eqref{eq:fhn} along with initial conditions given in \citep{chen2021physics} to learn the solutions of the system. 
% Similar to previous case, we assume that we have scarce data for $u$ available and that we know the physics for the evolution of $v$. Using these two available information, we train MUSIC by considering the setup used in \citep{chen2021physics}, also described below for reference.
% The ground truth data is generated using finite difference method with spatial timestep $dx = dy = 0.5$ and temporal timestep $dt = 0.0002$ for $t\in [0, 36]$ s. The ground truth data is then sampled over the temporal domain so the shape of the dataset for each $u$ and $v$ is $301\times 301\times 489$ with $t\in [7.18,36]$ s (see \citep{chen2021physics} for more details).
% We further downsample the dataset by uniformly projecting the data to a lower resolution spatio-temporal grid so the size of each variable $u$ and $v$ is $101\times 101\times 49$ (sample every third point in spatial domain and every tenth point in temporal domain). 

\subsection{$\lambda$-$\omega$ reaction diffusion system}
We consider a $\lambda-\omega$ reaction diffusion (RD) system in a 2D domain with the spiral pattern forming behavior governed by two coupled PDEs. Suppose $u$ and $v$ are two field variables, then the system of equations are given by
\begin{align*}
u_t &= 0.1 u_{xx} + 0.1 u_{yy} - u v^2 - u^3 + v^3 + u^2 v + u, \\
v_t &= 0.1 v_{xx} + 0.1 v_{yy} - u v^2 - u^3 - v^3 - u^2 v + v.
\end{align*}
The RD system can be used to describe wide range of behaviors including wave-like
phenomena and self-organized patterns found in chemical and biological systems \citep{chen2021physics}. Some applications can be found in pattern formation, ecological invasions, etc. 
% The coupled system is also referred to as an activator-inhibitor system because one state variable encourages the increase of both states while the other state component inhibits their growth. 
This particular RD equations in this test example displays spiral waves subject to periodic boundary conditions.
In order to generate the simulated training dataset, we solve the model using a finite difference method with inputs $x,y\in [-10,10]$ and $t\in [0,10]$. 
The initial conditions are selected as given in \citep{chen2021physics} so that the solution displays spiral behavior.
We downsample from a full solution with by uniformly selecting $128\times 128$ spatial points and 101 temporal steps.
Thus the dataset is of size $128\times128\times101$.
Since both the equations are similar, we randomly select $u$ as the equation variable and $v$ as the data variable.
Since both the solutions $u,v\in [-1,1]$, we do not use normalization.
Let $N_{s_x}$, $N_{s_y}$ and $N_t$ denote the number of spatial points in $x$ direction, spatial points in $y$ direction, and temporal points, respectively used for training and $N_{ic}$ denote the number of spatial data available at $t=0$. For each $i=1,\dots, N_{s_x}$, $j=1,\dots, N_{s_y}$ and $k = 1,\dots,N_t$, suppose $\widehat{u}_{ijk}$ and $(\widehat{v})_{ijk}$ denote the learned solutions at $(x_i,y_j,t_k)$, then loss function for optimizing the weights are given by 
\begin{align}
    \text{Loss} &= \dfrac{1}{N_{s_x}N_{s_y}N_t}\sum\limits_{i=1}^{N_{s_x}}\sum\limits_{j=1}^{N_{s_y}}\sum\limits_{k=1}^{N_t}
    \bigg(\left\|v_{ijk} - \widehat{v}_{ijk}\right\|_2^2\\
    &+ \left\| (\widehat{u}_t)_{ijk} - 0.1(\widehat{u}_{xx})_{ijk}
-0.1(\widehat{u}_{yy})_{ijk} + \widehat{u}_{ijk}\widehat{v}^2_{ijk} + \widehat{u}^3_{ijk} - \widehat{v}^3_{ijk} - \widehat{u}^2_{ijk}\widehat{v}_{ijk} - \widehat{u}_{ijk}\right\|_2^2 \bigg)\nonumber \\
&+ \dfrac{1}{N_{ic}}\sum\limits_{i=1}^{N_{ic}}\left\|u_i - \widehat{u}_i\right\|_2^2 + \lambda_0\sum\limits_{j=1}^{|\text{Neurons}|}\sigma(\alpha_j).\nonumber
\end{align}
 We use a fully connected neural network with a 80-20 training-validation split for model training. The ADAM optimizer is used with the learning rate tuned between $10^{-2}$ and $10^{-4}$ and $\lambda_0$ tuned between $10^{-2}$ to $10^{-12}$.

\begin{figure}[h!]
     \centering
     \begin{subfigure}[b]{1.0\textwidth}
         \centering
         \includegraphics[width=\textwidth]{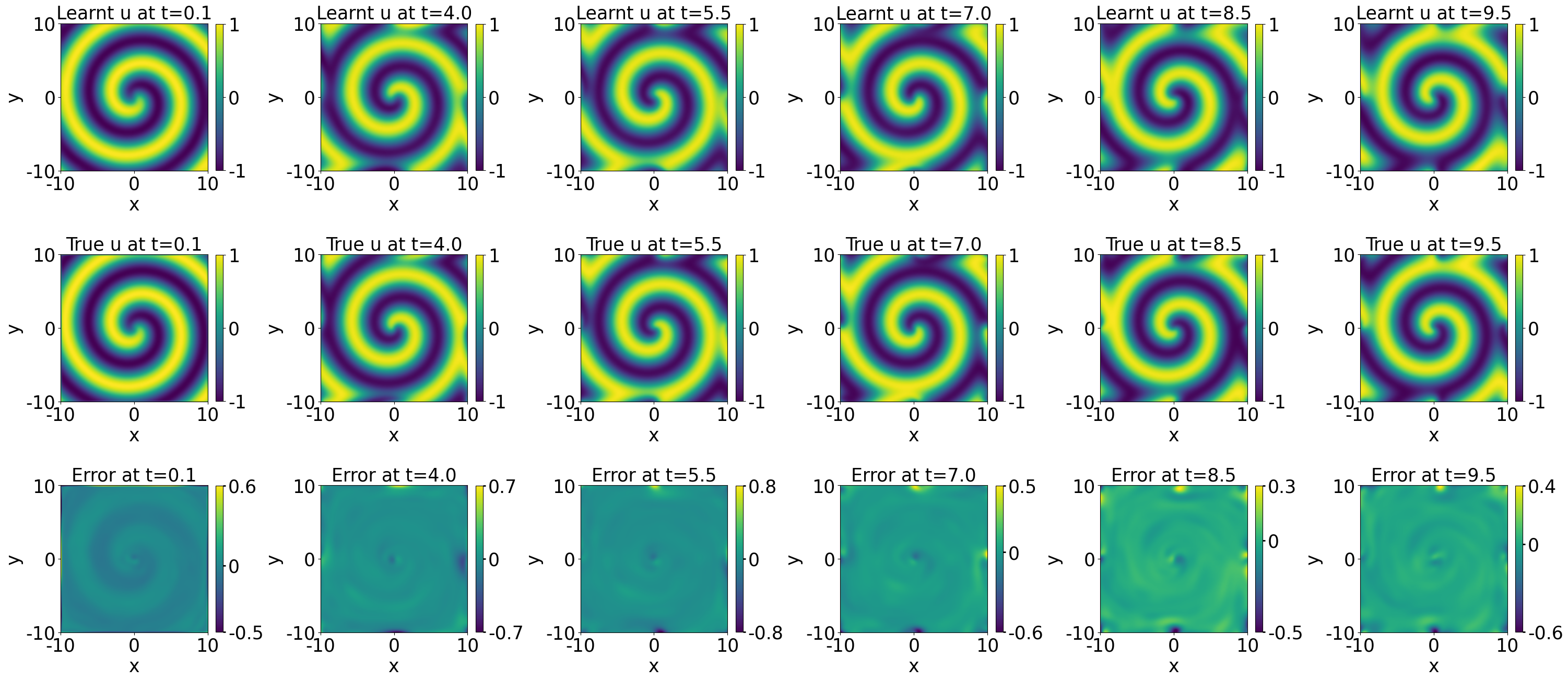}
         \caption{Predicted solutions (top row), true solution (middle row) and error (bottom row) for $u$.}
         \label{fig:rd_u}
     \end{subfigure}
     % \vspace{-5mm}
     % \hfill
     \begin{subfigure}[b]{1.0\textwidth}
         \centering
         \includegraphics[width=\textwidth]{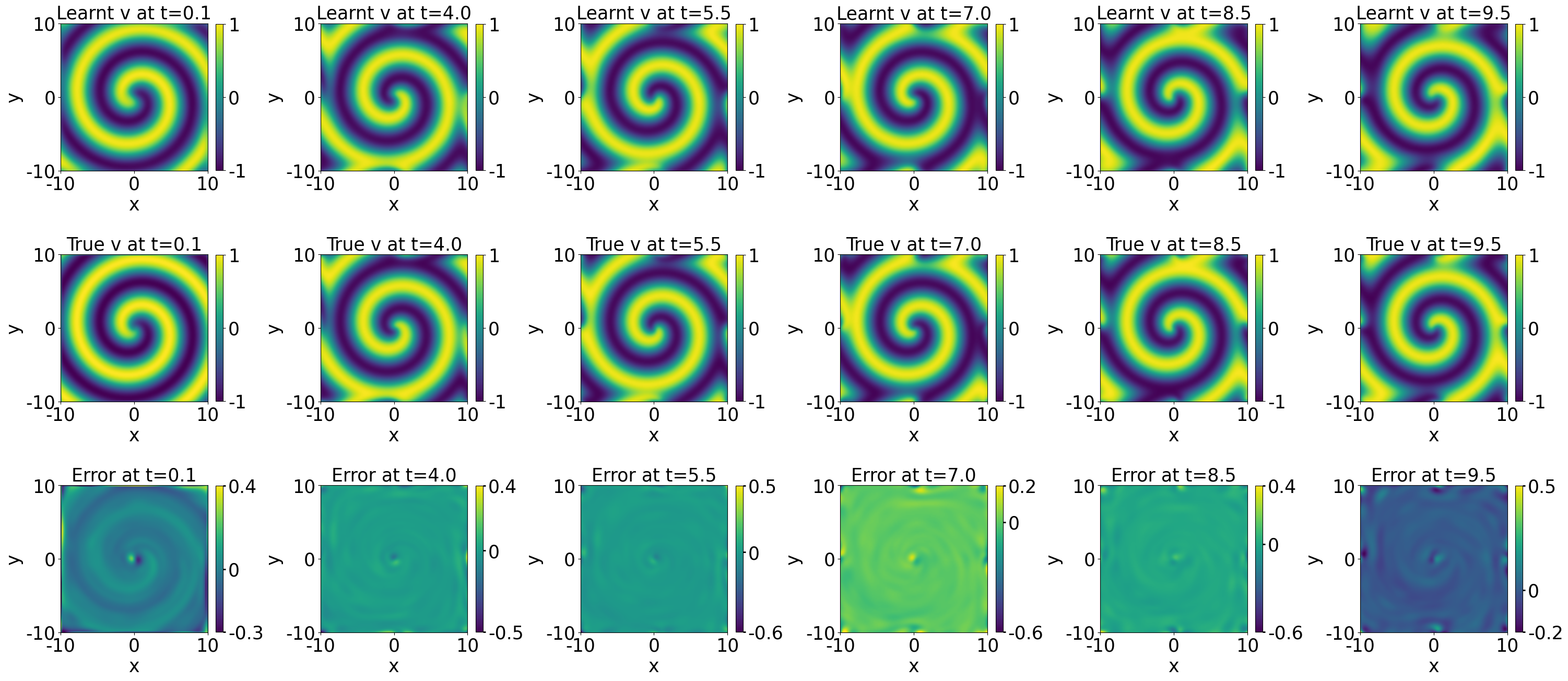}
         \caption{Predicted solutions (top row), true solution (middle row) and error (bottom row) for $v$.}
         \label{fig:rd_v}
     \end{subfigure}
        \caption{Solution to the RD system with random initialization learned using a 6 hidden layer 100 hidden dimension with $N_{s_x}*N_{s_y} = 5000$ and $N_t = 100$.}
        \label{fig:rd_uv}
\end{figure}
In the results demonstrated in Figure \ref{fig:rd_uv}, we see that for both the variables $u$ and $v$, the proposed model predicts the solutions which are visually indistinguishable from the true solutions. 
Upon plotting the error plots i.e. $u - \hat{u}$ and $v-\hat{v}$ ($u$, $v$ denotes the true solution and $\hat{u}$, $\hat{v}$ denotes the learned solution) we find the majority of the errors to be close to zero, with errors in some points close to the boundary of the spatial domain exceeding $|0.5|$. 
\begin{table}[ht!]
    \centering
    \begin{tabular}{|c|c|c|}
    \hline
        \backslashbox{Neurons}{Layers} & 4 & 6 \\
        \hline
        50 &  9.21\% $\pm$ 1.37\% & 8.92 \% $\pm$ 2.48\% \\
        Sparsity (\%) & 3.26\% & 2.77\%\\
        \hline
        100 &  8.26\% $\pm$  0.89\% & 8.03\% $\pm$ 0.93\% \\
       Sparsity (\%) & 1.65\% & 1.39\% \\
        \hline
    \end{tabular}
    \caption{Full-field average relative $\ell_2$ error (in percentage) for RD system. The model was trained using $N_{s_x}*N_{s_y} = 5000$, $N_t = 100$.}
    \label{tab:rd_neurons_layers}
\end{table}
The spatial boundary demonstrates higher error as we do not include any boundary conditions explicitly during model training. 
Also note that for this system as both $u$ and $v$ lie in $[-1,1]$ we omit normalization of inputs and outputs, thus making the bounds for $u - \hat{u}$ and $v-\hat{v}$ higher than previous examples.
Table \ref{tab:rd_neurons_layers} shows the performance of the model based on the average full-field $\ell_2$ error for different number of layers and neurons. 
A clear improvement is seen in model performance as the number of layers and neurons increases.
Not only does the error goes down, the confidence of interval also shrinks indicating a more stable model learning.
The sparsity percentage also goes down with an increase in model complexity.

\subsubsection{Effect of Noise and Timesteps on Learning Solutions}
In this section, we analyze the effects of noise as well as the number of timesteps used in the training of model.
% Both these cases offer an insight into how the model will perform should it be used on real world scenarios where the data collected may be noisy, or be temporaly scarce.
\begin{table}[ht!]
    \centering
    \begin{tabular}{|c|c|c|c|}
    \hline
        \backslashbox{$N_{s_x}*N_{s_y}$}{Noise (\%)} & 5\% & 10\% & 20\%\\
        \hline
        500 &  14.36\% $\pm$  2.11\% & 13.32\% $\pm$ 1.74\% & 12.26\% $\pm$ 3.72\% \\
        \hline
        1000 &  8.55\% $\pm$  0.75\% &  9.29\% $\pm$ 3.33\% & 11.83\% $\pm$  2.1\%\\
        \hline
    \end{tabular}
    \caption{Full-field average relative $\ell_2$ error (in percentage) for RD system with different levels of noise and $N_{s_x}*N_{s_y}$. The model was trained using $N_t=100$, 6 layers and 100 neurons.}
    \label{tab:rd_ns-noise}
\end{table}
We train a model with 6 layers and 100 neurons with noisy dataset. 
The number of spatial points is varied to get an idea of how the noise level affects the prediction errors based on available data.
In Table \ref{tab:rd_ns-noise} we see that when the number of spatial points used in training is less at $N_{s_x}*N_{s_y}=500$, errors for all noise levels are over 10\%. 
Increasing $N_{s_x}*N_{s_y}=1000$ improves model learning significantly and the relative errors drop to below 10\% for noise levels 5\% and 10\%. 
For a higher noise level of 20\%, while the error remains over 10\% for any value of $N_{s_x}*N_{s_y}$, a marginal improvement is observed upon increasing $N_{s_x}*N_{s_y}$ from 500 to 1000. 

\begin{figure}[ht!]
    \centering
\includegraphics[width=0.5\linewidth]{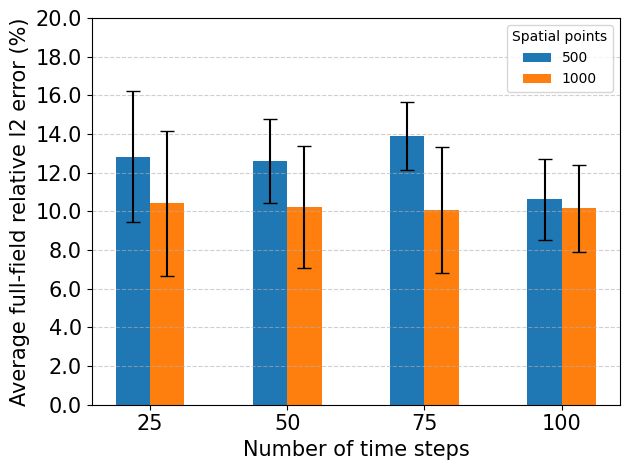}
    \caption{Average relative $\ell_2$ error for varying $N_t$ in FN system. Model uses 6 layers, 100 neurons.}
    \label{fig:rd-timesteps-ns}
\end{figure}

Further analysis into the number of timesteps indicated that irrespective of the $N_t$ used, the $\ell_2$ errors do not vary much. 
Infact, to achieve a performance level similar to that of 100 timesteps, all $N_t=\{25,50,75\}$ give similar errors (Figure \ref{fig:rd-timesteps-ns}).
However, increasing the number of spatial points from $N_{s_x}*N_{s_y}=500$ to $N_{s_x}*N_{s_y}=1000$ improves the errors by almost 2\%.
Further increasing $N_{s_x}*N_{s_y}$ to 5000 could lead to lower errors (see Table \ref{tab:rd_neurons_layers}).

\subsection{Physical Model of Wildfires}\label{sec:physical-wildfire}
Physical models of wildfire are built from conservation laws which model the rate of change of fire temperature based on conservation of energy, balance of fuel supply and fuel reaction rate. The coupled system is reaction-diffusion like system, has been described in multiple works \citep{mandel2008wildland,san20232d,8966412} and is given as,

\begin{equation}\label{eq:wf}
\begin{aligned}
    u_t &= \kappa\Delta u - \mathbf{v}\cdot\nabla u + f(u,\beta), \,\, &&\text{ in } \Omega \times (0,t_{max}]\\
    \beta_t & = g(u,\beta), \,\, &&\text{ in } \Omega \times (0,t_{max}]\\
    % u(\mathbf{x},t) & = u_{\Gamma}(\mathbf{x},t), \,\, && \text{on } \Gamma \times (0,t_{max}]\\ 
    % \beta(\mathbf{x},t) & = \beta_{\Gamma}(\mathbf{x},t), \,\, && \text{on } \Gamma \times (0,t_{max}]\\
    % u(\mathbf{x},0) & = u_0(\mathbf{x}), \,\, && \text{in } \Omega\\ 
    % \beta(\mathbf{x},0) & = \beta_0(\mathbf{x}), \,\, && \text{in } \Omega 
 \end{aligned}
 \end{equation}
 where $u$ and $\beta$ denote the firefront temperature and fuel availability respectively, $t_{max}$ denotes the final time, $\mathbf{v}$ denotes the vector field of wind and topography. The the definition of other constants and the values used are given in Table \ref{tab:symbols}. The functions $f$ and $g$ can be described in various ways. A common representation used in prior works \citep{san20232d} is defined as, 
 \begin{equation}\label{eq:wf_fg}
 \begin{aligned}
     f(u,\beta) &= H_{pc}(u)\beta\exp\left(\dfrac{u}{1+\epsilon u}\right) - \alpha u,\\
     g(u,\beta) &= -H_{pc}(u)\dfrac{\epsilon}{q}\beta\exp\left(\dfrac{u}{1+\epsilon u}\right),\\
     H_{pc}(u) & =  \begin{cases} 
      1, & \text{if } u\geq u_{pc} \\
      0, & \text{otherwise} 
   \end{cases}
 \end{aligned}
 \end{equation}
Assuming that the spatial domain is large enough to avoid fire spreading at the boundary $\partial\Omega$, we assume Dirichlet boundary conditions $u_{\partial\Omega}(\mathbf{x},t) =\beta_{\partial\Omega}(\mathbf{x},t) = 0$ on $\partial\Omega\times (0,t_{max}].$
This model can be solved numerically to obtain fire spread dynamics in a given spatio-temporal domain.

\begin{figure}[h!]
\centering
\begin{subfigure}[b]{0.3\textwidth}
    \centering
    \includegraphics[scale = 0.2]{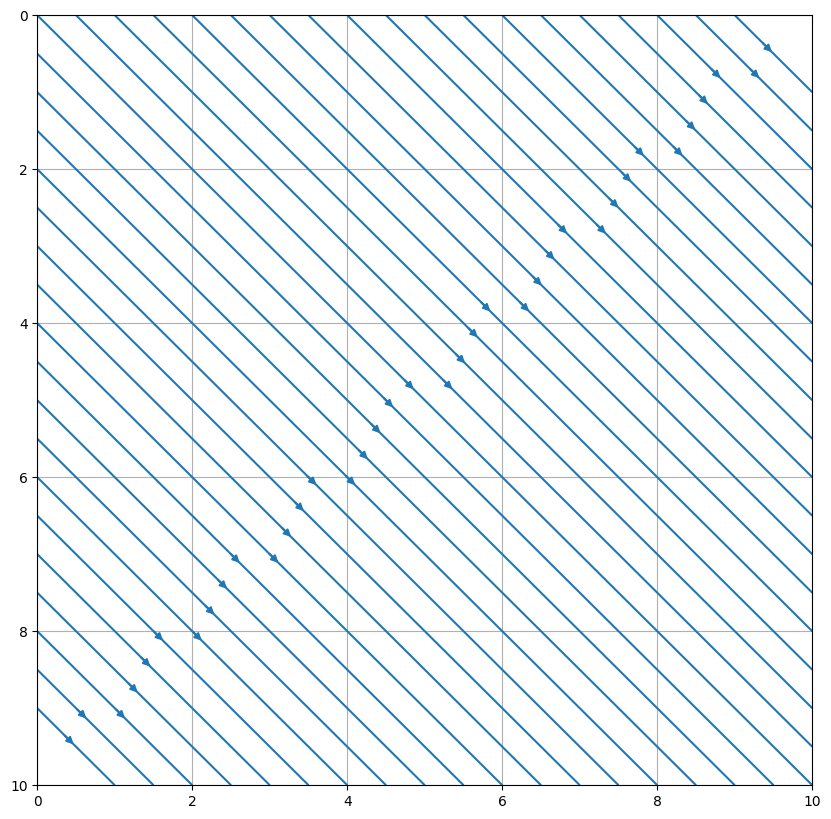}
    \caption{Fixed wind vector}
    \end{subfigure}
    \centering
\begin{subfigure}[b]{0.3\textwidth}
    \centering
    \includegraphics[scale = 0.2]{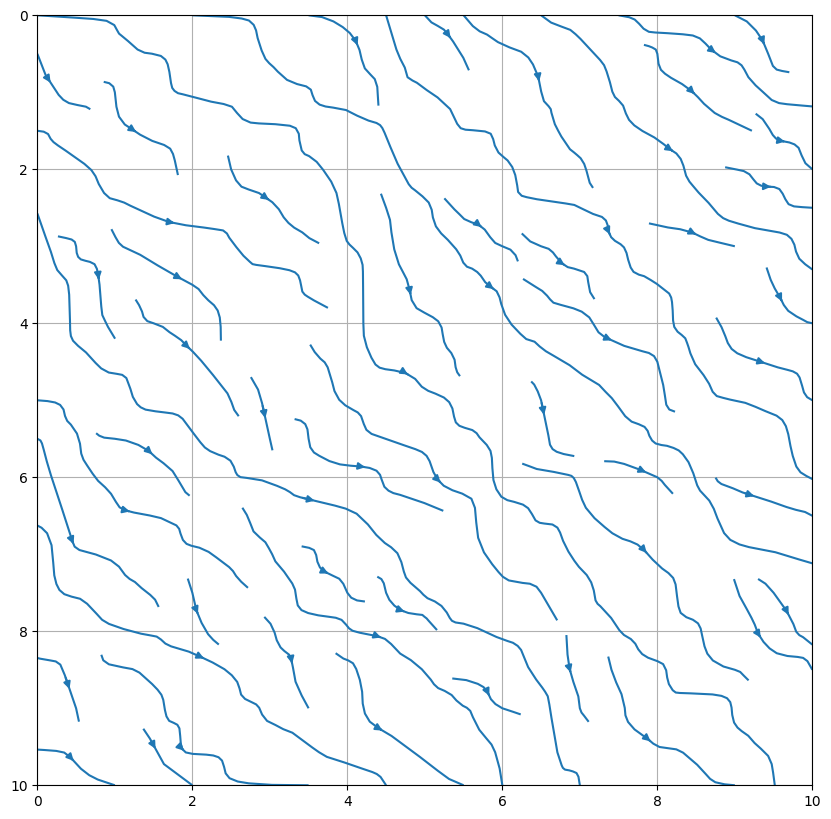}
    \caption{Wind vector $(u^2,v^2)$.}
    \end{subfigure}
    % \begin{subfigure}[b]{0.3\textwidth}
    % \centering
    % \includegraphics[scale = 0.2]{windmaps/RandomWind.png}
    % \caption{Wind vector $(u,v)$}
    % \end{subfigure}
    \caption{Wind field plots used to solve the physical model.}
    \label{fig:wind}
\end{figure}

 \begin{table}[h!]
     \centering
     \begin{tabular}{|c|c|c|}
     \hline
     Symbol & Description & Values \\
     \hline
     $\Omega$ & Spatial domain & $[0,10]\times [0,10]$\\
     $t_{max}$ & Final time & 10 \\
          $\kappa$ & diffusion coefficient & 0.2 \\
          $\epsilon$ & Inverse of activation energy of fuel& 0.3\\
          $\alpha$ & Natural convection & 0.01 \\
          $q$ & Reaction heat & 1 \\
          $u_{pc}$ & Phase change threshold & 3\\
          \hline
     \end{tabular}
     \caption{Description of various parameters used in the physical model of wildfire.}
     \label{tab:symbols}
 \end{table}
% We define the spatial coordinates as a normalized 2D plane with coordinates $(x,y)\in [0,1]\times [0,1]$. 
We select the data variable $u$ and equation variable $\beta$, as it is much easier to understand the dynamics of how the fire spread burns the vegetation than the physics of $u$.
Defining the equation for $u_t$ is much more complex due to the interaction of multiple variables such as wind, moisture, etc. 
Training data is obtained by solving the system (equations \eqref{eq:wf} and \eqref{eq:wf_fg}) with $\Delta x = \Delta  y=0.2$ and $\Delta t=0.1$. 
The environmental parameters present in the physical model (vegetation, wind vectors) are either simulated using known function(s) (for vegetation) or taken from prior works \citep{san20232d}.
We also discard any data for $t\in [0,2]$ as observed data is often available (captured by satellites or other measurement technology) only after the firefront temperature reaches a certain magnitude which occurs after a certain number of timesteps. 
Our selection of the interval $[0,2]$ is based on threshold value after which the fuel availability will start going down (corresponds to burned regions). 
The final full resolution dataset is of size $51\times 51\times 80$. 
Since the solution $u\in [0,10]$ and $b\in[0,1]$ demonstrated varying range of outputs, we normalized all the input and output data using the min-max normalization so that all values are within 0 and 1.
 To make the simulations more realistic, we consider different $\mathbf{v}$ vectors: linear (fixed) $\mathbf{v}$, and a combination of random wind fields $\mathbf{v} = (v_1^2,v_2^2)$ where $v_1,v_2\sim \mathcal{N}(0,0.5)$. 
 A sample of the wind field plots used for solving the physical model are given in Figure \ref{fig:wind}.
Let $N_{s_x}$, $N_{s_y}$ and $N_t$ denote the number of spatial points in $x$ direction, spatial points in $y$ direction, and temporal points, respectively used for training and $N_{ic}$ denote the number of spatial data available at $t=0$ (corresponds to  $t=2$ before data was normalized). 
If a model $h_{\Theta}(\mathbf{x},t)$ is trained to learn the solutions, such that for each $i=1,\dots, N_{s_x}$, $j=1,\dots, N_{s_y}$ and $k = 1,\dots,N_t$, suppose $\widehat{u}_{ijk}$ and $(\widehat{b})_{ijk}$ denote the learned solutions at $(x_i,y_j,t_k)$, then loss function for optimizing the weights with sparsity level $k$ is given by 
\begin{align}
    \text{Loss} &= \dfrac{1}{N_{s_x}N_{s_y}N_t}\sum\limits_{i=1}^{N_{s_x}}\sum\limits_{j=1}^{N_{s_y}}\sum\limits_{k=1}^{N_t}
    \bigg(\left\|u_{ijk} - \widehat{u}_{ijk}\right\|_2^2+ \left\| (\widehat{b}_t)_{ijk} - g(\widehat{u}_{ijk},\widehat{b}_{ijk})\right\|_2^2 \bigg) + \dfrac{1}{N_{ic}}\sum\limits_{i=1}^{N_{ic}}\left\|b_i - \widehat{b}_i\right\|_2^2\nonumber\\
     \Theta & = \text{Top}_k (\Theta),
\end{align}
where Top$_k (\Theta)$ denotes the thresholding function that keeps only top $k$ parameters by magnitude and zeros out the rest. 
The final trained model only contains $k$ non-zero weights (and biases).
Similar to previous examples, we use a 80-20 training and validation split for MUSIC training with a ReLU activation function which ensures that all the outputs remain positive as required by the solutions.
Moreover, the ReLU activation is a better choice for handling discontinuities in the data and model.
The ADAM optimizer is used to optimize the parameters with the learning rate tuned between $10^{-3}$ and $10^{5}$.

\subsubsection*{Using Hard Thresholding Instead of Gated $\ell_0$ for Physical system of Wildfire}
For inducing sparsity in this system, we use an iterative thresholding method, where for each epoch the model only keeps the top $k$ weights (and biases) by magnitude and zeros out the rest.
We avoid using gated $\ell_0$ for several reasons. 
Post normalization, the data variable  $u$ is majorly dominated by zeros. 
Thus, for a model whose majority of the training data is characterized by zero values, it is highly likely that using sparsity promoting method like gated $\ell_0$ will lead to model collapse/ unstable training.
In particular, even if the model learns a zero solution, the errors would still be below the convergence threshold due to data imbalance. 
Thus, to solve this issue, we could either reduce the interval length of the spatial domain so that only the patch around the burning region is considered. In this section, we instead choose to use a different approach where we use weight thresholding during training process. 
% Instead of thresholding the weights after every epoch, we let the model train for certain number of iterations and then apply hard thresholding \citep{blumensath2009iterative,foucart2011hard} to zero out weights that are closer to zero.

% \subsubsection{Single Fire}
\subsubsection{Fixed Wind Vector}
In this section, we discuss in detail the model performance when the fire spread is driven by a fixed vector $\mathbf{v}$. 
\begin{figure}[h!]
     \centering
     \begin{subfigure}[b]{1.0\textwidth}
         \centering
         \includegraphics[width=\textwidth]{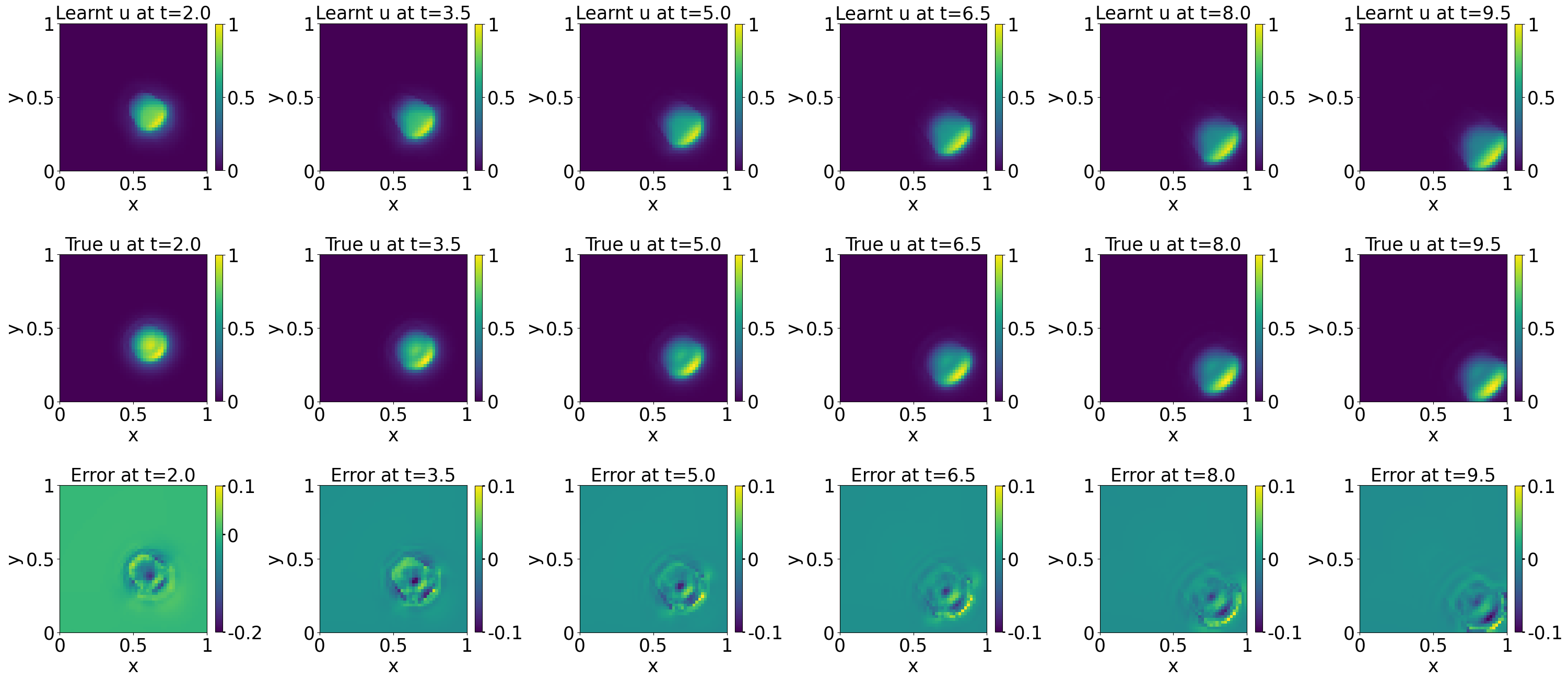}
         \caption{Predicted solutions (top row), true solution (middle row) and error (bottom row) for $u$.}
         \label{fig:wildfire_u}
     \end{subfigure}
     % \vspace{-5mm}
     % \hfill
     \begin{subfigure}[b]{1.0\textwidth}
         \centering
         \includegraphics[width=\textwidth]{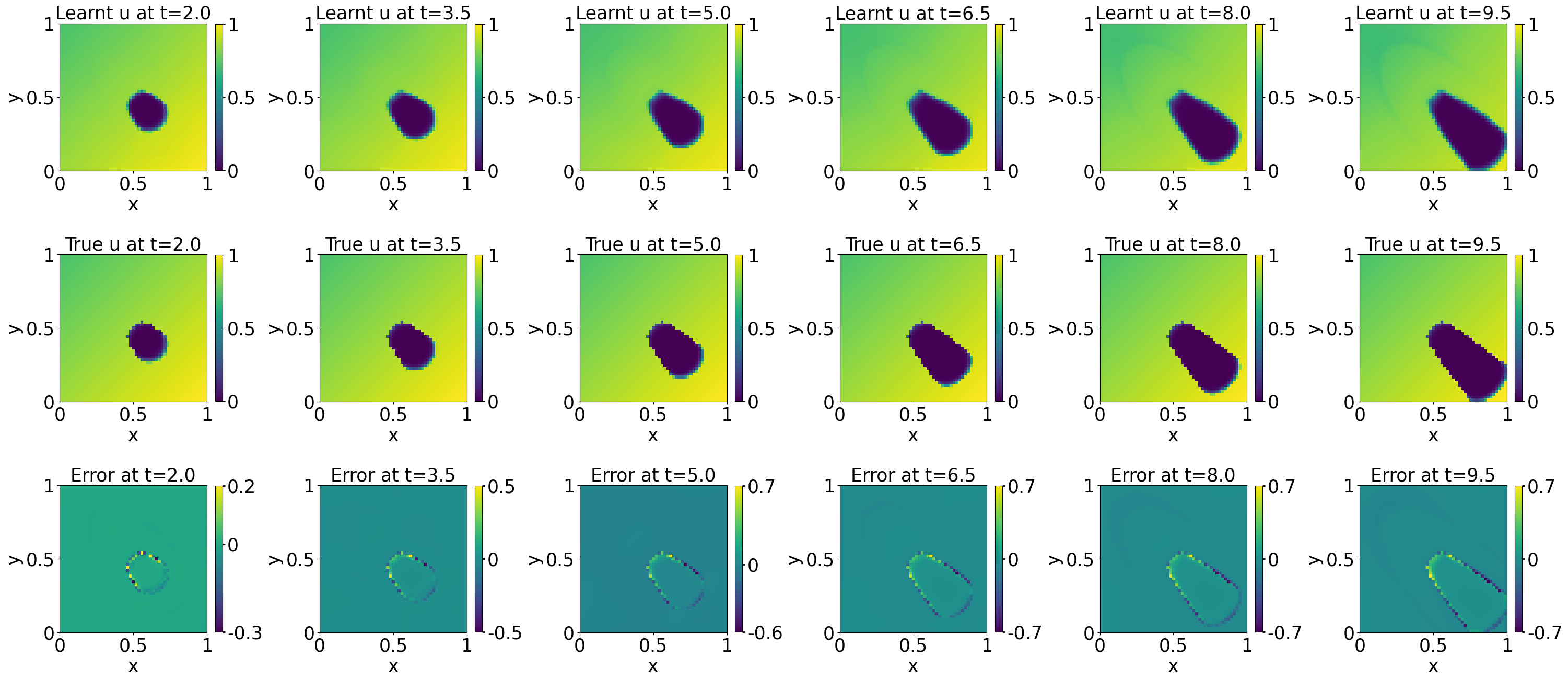}
         \caption{Predicted solutions (top row), true solution (middle row) and error (bottom row) for $\beta$.}
         \label{fig:wildfire_b}
     \end{subfigure}
        \caption{Solution to the physical model of wildfire with \textbf{fixed wind vector} learned using a 6 layer fully connected neural networks using 100 neurons with $N_{s_x}*N_{s_y} = 1000$ and $N_t = 60$.}
        \label{fig:wildfire_ub}
\end{figure}
The full-field solutions generated by MUSIC along with the error plots are given in Figure \ref{fig:wildfire_ub}.
For predicting the fire temperature $u$, the (normalized) predicted values lie very close to the true vales. The error $u-\hat{u}$ always remains below 0.2 across all the timesteps with some higher errors seen either at the fire perimeter or in regions where the fire temperature varies rapidly.
For predicting the burned points in $\beta$, note that while for most of the regions the solution is accurate, the error jumps at the perimeter of the burned area. 
This happens because the system involves discontinuous functions $
f$ and $g$ leading to the true solution also having sudden jumps.
The universal approximation theorem states that a neural network can learn any continuous function. However, as our solutions are discontinuous, the model learns a close continuous approximation to the true solution, thus leading to a significantly higher error at the perimeter.

\begin{table}[h!]
    \centering
    \begin{tabular}{|c|c|c|}
    \hline
        \backslashbox{Neurons}{Layers} & 4 & 6 \\
        \hline
        50 &  7.403 \% $\pm$ 0.76\% & 6.37 \% $\pm$ 1.13\% \\
        Sparsity (\%)& 50\% & 50\%\\
        \hline
        100 & 7.08\% $\pm$ 1.44 \% & 5.74\% $\pm$ 0.30\% \\
        Sparsity (\%) & 50\% & 50\% \\
        \hline
    \end{tabular}
    \caption{Full-field average relative $\ell_2$ error (in percentage). The model was trained using $N_{s_x}*N_{s_y} = 1000$, $N_t = 60$.}
    \label{tab:wildfire_neurons_layers}
\end{table}
In Table \ref{tab:wildfire_neurons_layers} we report the full-field average relative $\ell_2$ error for varying neural network layers and neurons. We find that as the number of trainbale parameters increase, the accuracy of the model also increases with lower error and lower variance (standard deviation). 
\textcolor{black}{We also run the model with different values of $N_t$ to test the effect of number of timesteps model training and note that $N_t$ does not majorly affect $\ell_2$ error.
Although, if enough number of spatio-temporal points are used ($N_t=1000$ and $N_t$=80), then we get the lowest error. 
}
\begin{figure}[h!]
     \centering
     \begin{subfigure}[b]{0.45\textwidth}
         \centering
         \includegraphics[width=\textwidth]{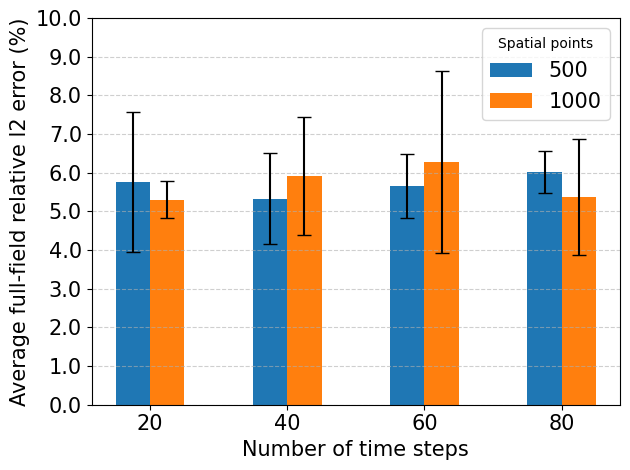}
         \caption{Full relative $\ell_2$ error for $N_t$ versus $N_s$. Model had 6 layers and 50 neurons.}
         \label{fig:wildfire-nt-vary}
     \end{subfigure}
     % \vspace{-5mm}
     % \hfill
     \begin{subfigure}[b]{0.5\textwidth}
         \centering
         \includegraphics[width=\textwidth]{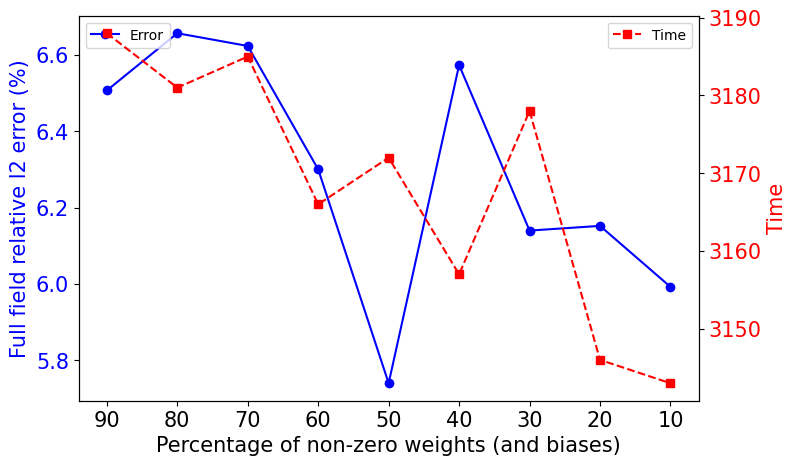}
         \caption{Full relative $\ell_2$ errors and training time for different weight sparsity. Model uses $N_t=60$ and $N_s=1000$, 6 layers and 100 neurons.}
         \label{fig:wildfire-sparsity-vs-time}
     \end{subfigure}
        \caption{Effect of $N_t$, $N_s$ and varying sparsity in model training.}
        \label{fig:wildfire-sparsity-time-vary-nt-full}
\end{figure}
In Figure \ref{fig:wildfire-sparsity-vs-time} it can be clearly seen that as the percentage of nonzero parameters go down (increasing sparsity), the error improves, with optimal performance at 50\% sparsity. 
The training time also demonstrates a downward trend, although marginally.
However, for larger datasets and complex models, the reduction in training time for higher sparsity will be more evident and significant.

% \begin{table}[h!]
%     \centering
%     \begin{tabular}{|c|c|c|c|c|}
%     \hline
%         \backslashbox{$N_s$}{$N_t$} & 20 & 40 & 60 & 80  \\
%          \hline
%          500 & 5.76\% $\pm$ 1.82\% & 5.33\%$\pm$ 1.17\% &
% 5.66\% $\pm$ 0.82\% & 6.01\% $\pm$ 0.54\% \\
%         1000 & 5.30\% $\pm$ 0.48\% & 5.91\%$\pm$ 1.52\% &
% 6.28\% $\pm$ 2.35\% & 5.37\% $\pm$ 1.50\% \\
% \hline
%     \end{tabular}
%     \caption{Full-field average $\ell_2$ error (in percentage) for varying $N_t$. The model was trained using $N_{s_x}*N_{s_y} = 1000$, 6 layers and 100 neurons.}
%     \label{tab:wildfire-varynt}
% \end{table}

\subsubsection{Stochastic Wind and Topography Vector}
In reality, the vector $\mathbf{v}$ is often driven by stochasticity
In this section, we use data generated by a vector $\mathbf{v}=(v_1^2,v_2^2)$, where $v_1,v_2\sim\mathcal{N}(0,0.5)$.
We first use our formulation to learn the model solution when there the fire driven by the varying wind vector.
\begin{figure}[h!]
     \centering
     \begin{subfigure}[b]{1.0\textwidth}
         \centering
         \includegraphics[width=\textwidth]{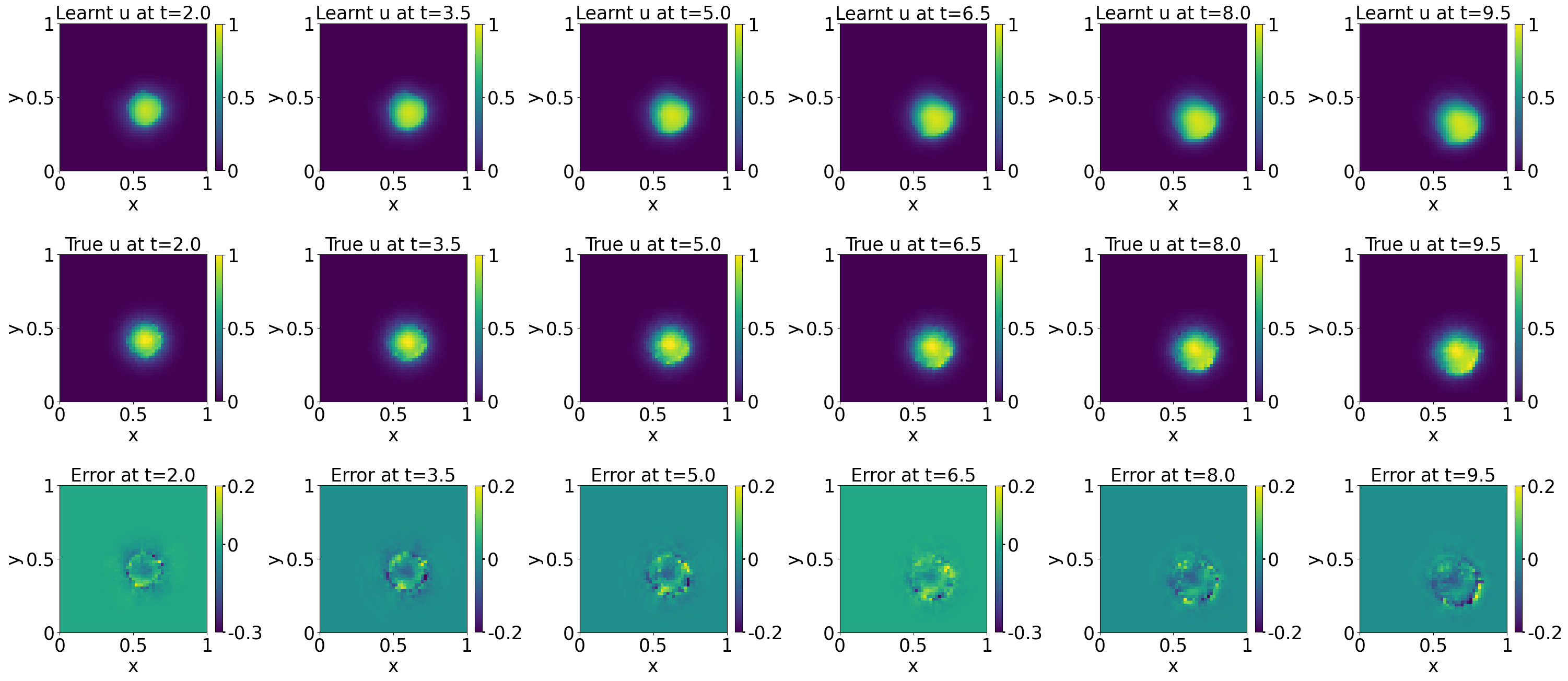}
         \caption{Predicted solutions (top row), true solution (middle row) and error (bottom row) for $u$.}
         \label{fig:wildfire-multi-windquad_u}
     \end{subfigure}
     % \vspace{-5mm}
     % \hfill
     \begin{subfigure}[b]{1.0\textwidth}
         \centering
         \includegraphics[width=\textwidth]{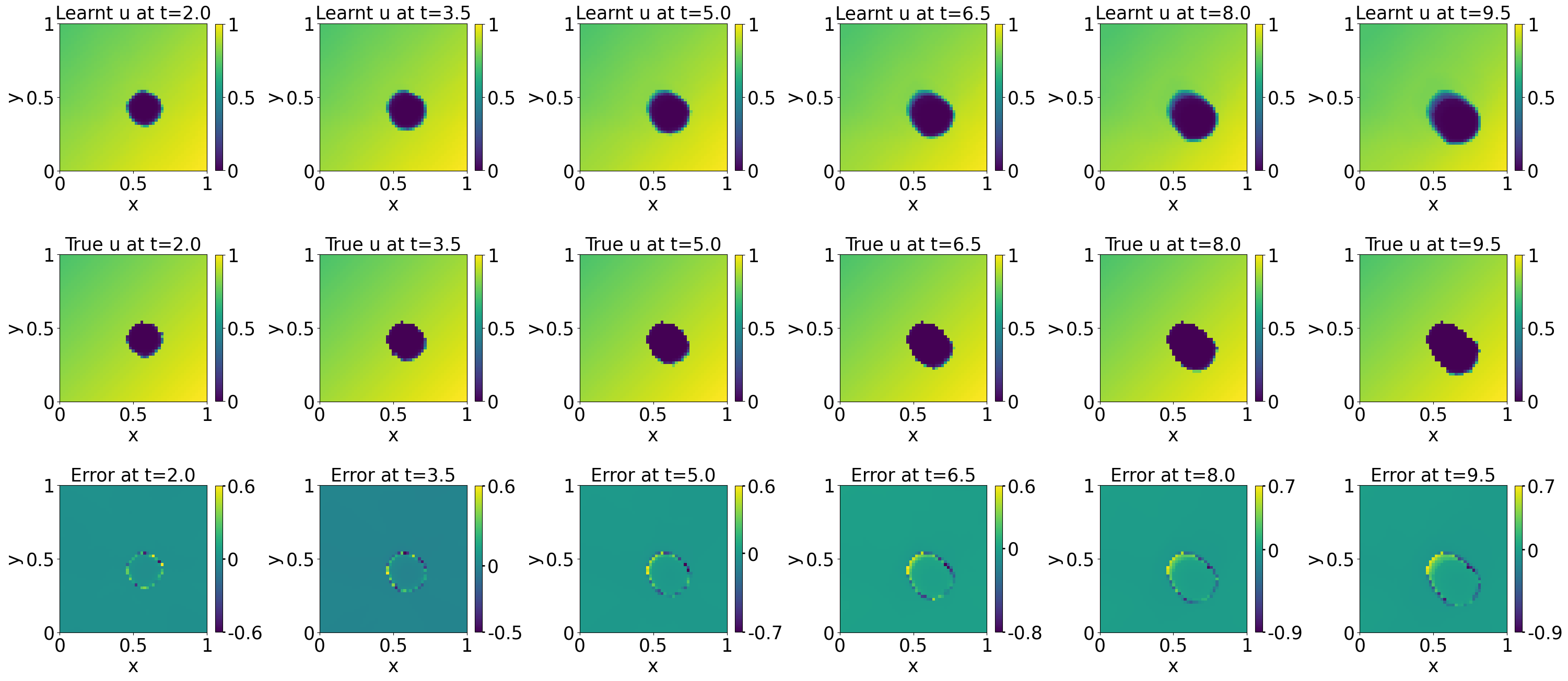}
         \caption{Predicted solutions (top row), true solution (middle row) and error (bottom row) for $\beta$.}
         \label{fig:wildfire-multi-windquad_b}
     \end{subfigure}
        \caption{Solution to the physical model of wildfire with \textbf{varying wind vector and single initial fire} learned using a 6 layer fully connected neural networks using 100 neurons with $N_{s_x}*N_{s_y} = 1000$ and $N_t = 60$. The average relative $\ell_2$ error is $5.82 \% \pm 0.23\%$ with 50\% model sparsity.}
        \label{fig:wildfire-single-windquad_ub}
\end{figure}
The average full-field $\ell_2$ error is $5.82\%$ with a variation of about $0.23\%$ indicating the possibility learning fire spread dynamics in a realistic setting where the parameters are driven by stochasticity. 
Figure \ref{fig:wildfire-single-windquad_ub} shows under varying wind conditions, the trained model learns the spread of fire which is less smooth and erratic in comparison to a spread driven by fixed $\mathbf{v}$.
% The fire spread in this particular case is relatively slower.
For both $u$ and $\beta$, MUSIC captures this irregularity.
As pointed out previously, the errors around the perimeter of fire $u$ and burned vegetation $\beta$ are more as the neural network learns a continuous approximation of the true solution (which is clearly discontinuous).

\section{Discussion}
In this section, we discuss various aspects of the proposed model in terms of its convergence history, comparison between both types of sparsity, effects of sparsity and model performance when forecasting solutions on an unseen temporal domain.
Note that we pick one system and do a detailed analysis for each of the aspects mentioned above. 
For most cases, we use the SWE system as it is computationally quick and easier to analyze and visualize than the other systems. 
Analysis and conclusions drawn for all the other systems will be similar to the ones mentioned below.

\subsection{Why we include initial conditions but not boundary conditions in training?}
Prior works \citep{raissi2019physics} have suggested the need to incorporate boundary conditions explicitly to ensure that the model is trained to satisfy the physics at the spatial boundary.
In our proposed model we do not include any loss terms with respect to the boundary conditions. 
Regardless, MUSIC can learn the true solutions, including for majority of the spatial boundary. 
This is possible due to our mesh free (random) sampling of training data.
Note that the mesh free sampling of data (and collocation points) from the spatio-temporal domain also selects points from the spatial boundary.
This only holds true for spatial dimension greater than or equal to two.
However, for SWE system with one dimensional spatial domain, the boundary points (i.e., $x=0$ and $x=L$) may not necessarily get selected during mesh free sampling.
Thus, not including the boundary can lead to incorrect solution learning at $x=0$ and $x=L$, as can be seen from our plots in Figure \ref{fig:swe-sol-best}. 
However, as the solutions at the boundary is same as in the neighborhood of $x=0,L$, the error/deviation from the true solution is relatively small.
This can however cause significant errors for variables whose solutions inside the domain differ from the boundary.

The initial condition however needs to be enforced due to the same reason as explained above for the 1D spatial domain case. 
Since we sample from the temporal domain separately (which is only one dimension), the probability of $t=0$ being sampled is much lower. 
Moreover, without any knowledge of the true solution at $t=0$, the model could be learning any of the possible solutions to the system that preserve the enforced physics.

% \subsection{Swapped Variables}
% For RD
% For FHN or SWE

\subsection{Convergence history}
\begin{figure}[h!]
     \centering
     \begin{subfigure}[b]{0.4\textwidth}
         \centering
\includegraphics[width=\textwidth]{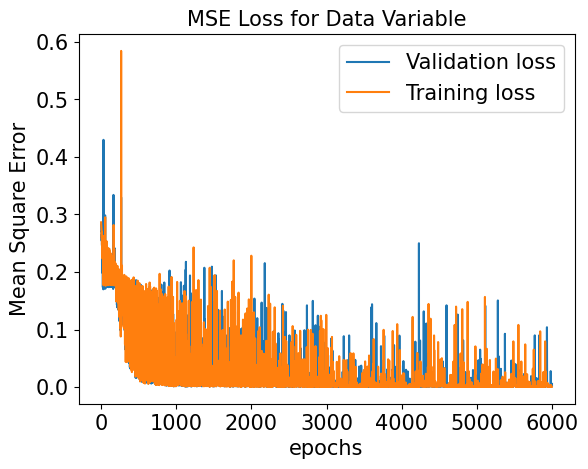}
         \caption{MSE for learning $hu$ using data.}
         \label{fig:mse-data}
     \end{subfigure}
        \begin{subfigure}[b]{0.4\textwidth}
         \centering
    \includegraphics[width=\textwidth]{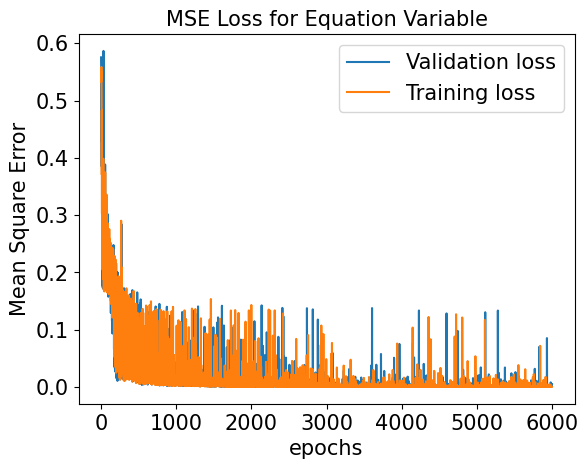}
         \caption{MSE for learning $h$ using equation.}
         \label{fig:mse-eq}
     \end{subfigure}
         \begin{subfigure}[b]{0.4\textwidth}
         \centering
         \includegraphics[width=\textwidth]{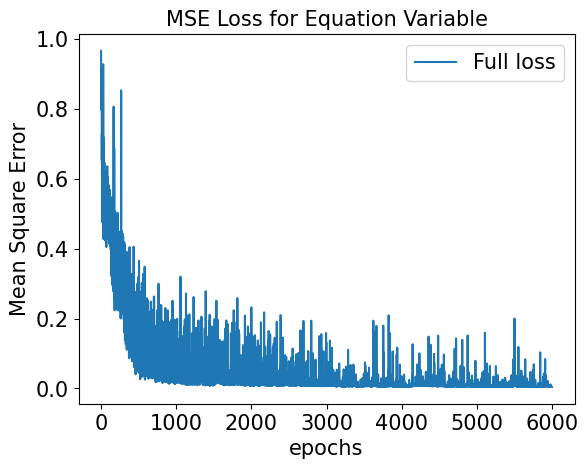}
         \caption{Overall MSE}
         \label{fig:mse-full}
     \end{subfigure}
         \begin{subfigure}[b]{0.4\textwidth}
         \centering
         \includegraphics[width=\textwidth]{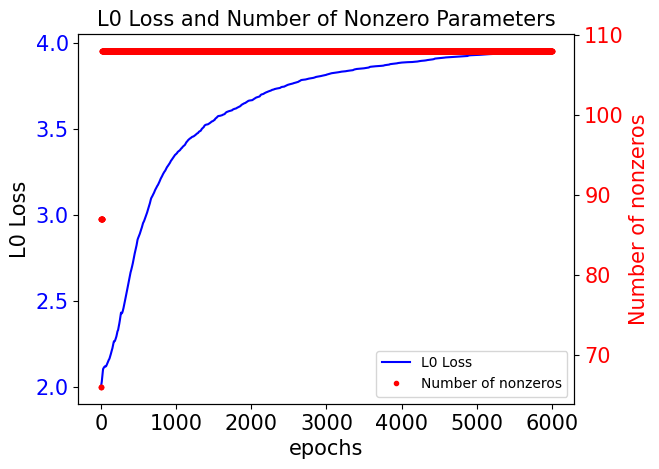}
         \caption{$\ell_0$ loss and number of non zeros.}
         \label{fig:mse-sparse}
     \end{subfigure}
        \caption{MSE for different terms and different variables for SWE system. Model was trained with 4 layers, 20 neurons, $N_s=100$ and $N_t$ = 800.}
        \label{fig:swe-convergence}
\end{figure}

In order to assess the convergence history of MUSIC, we pick the SWE model to assess how the model learns over the training domain, validation domain and finally forecasts solutions over the full domain.
In Figure \ref{fig:swe-convergence} we plot the mean square errors for on the training set and validation set for both $h$ and $hu$ for each epoch. Also plotted in Figure \ref{fig:mse-sparse} is the value of $\|\Theta\|_0$ approximated via a continuous distribution.

From Figures \ref{fig:mse-data}, \ref{fig:mse-eq} and \ref{fig:mse-full} we see that the MSE of the model starts to fall right from the beginning for both the equation and the data variable. 
The variance of the MSE is high for the first 2000 epochs (approximately) but later falls below 0.1 with the errors going very close to zero.
Even though we use disjoint priors (data variable $hu$ and equation variable $h$), the model architecture clearly facilitates the learning of both the variables.
The model initially starts with very high level of sparsity (Figure \ref{fig:mse-sparse}) and as training progresses, the $\ell_0$ loss term increases, leading to an increase in the number of non-zero parameters which eventually plateau out to about 110 parameters with $\ell_0$ loss value of $\sim $ 4.0.
The plot clearly indicates that even with sparsity constraints, the model gradually opens the gates as needed to ensure an improvement in the accuracy. 

\subsection{Structured and Unstructured Sparsity}
In this section we compare both neuron versus weight sparsity and briefly discuss the advantage/drawback of each model. 
We use SWE system as a benchmark to test both neuron and weight sparsity.
We trained the model by using a weight thresholding where every (few) iteration(s) only a certain number of weights were retained based on their magnitude and the rest were made zero.
Different levels of sparsity were used for training. For this case, we noticed that more number of nonzeros gave better results.
In Table \ref{tab:iht-vs-l0} we see that as the percentage of nonzeros go down, the full-field relative $\ell_2$ error for unstructured sparsity goes up. Upon comparing the same number of nonzeros (about 8\%) for unstructured versus structured sparsity, we see the clear difference in performance, where the error using structured sparsity if below 4\% while for unstructured sparsity it is above 70\%.
\begin{table}[h!]
    \centering
    \begin{tabular}{|c|c|c|c|c|}
    \hline
        Sparsity type & \multicolumn{3}{c|}
        {Unstructured (Weight $\ell_0$)} & Structured (Neuron $\ell_0$) \\
        \hline
        Number of nonzeros (\%) & 90 & 50 & 8 & 7.93\\ 
        Relative $\ell_2$ error & 8.14\% & 27.14\% & 73.83\% & 3.98\%\\
        \hline
    \end{tabular}
    \caption{SWE system solutions learned using structured and unstructured sparsity using 4 Layers, 20 neurons, $N_s=100$ and $N_t=800$.}
    \label{tab:iht-vs-l0}
\end{table}
Upon comparing the number of nonzeros, structured sparsity leads to a better model compression since it shuts off neurons leading to shutting down of all incoming and outgoing weights of that `off'. 
The trained compressed model can be thought of as the best sub-network that learns the solutions. 
In unstructured sparsity, all neurons may end up remaining active, as long as each neuron as atleast one non-zero incoming weight. 
Thus, the overall model may not be compressed fully as important weights of the model may be scattered all over.

However, note that there may be cases when unstructured sparsity can outperform structured sparsity (see \ref{sec:physical-wildfire}). When the dataset structure is dominated by a certain value (e.g. zeros in the wildfire system example), structured sparsity can lead to all neurons being turned `off' since even a model learning only zero/constant value will have training and validation loss below the convergence threshold.
In such cases, the unstructured sparsity can give better and reliable results while still inducing certain degree of sparsity and model compression.

\subsection{Predicting the Solution for Unseen Time Domain}\label{sec:swe-forecast}

In this section, we try to asses how much data in the temporal domain data is required for an accurate out of (time) domain (OOD) forecasting. We divide the time domain into to parts $[t_0,t_1]$ and $(t_1,t_2]$ where $t_0$ denotes the initial time (often 0 after normalization), $t_1$ denotes the time till which the solutions will be considered for training the model, and $t_2$ is the final time (often 1 after normalization). The model will be trained as before by considering randomly chosen spatio-temporal points from $[t_0,t_1]$ and then the predicted solutions on $(t_1,t_2]$ will be compared with the true solutions. The relative $\ell_2$ error for predicting $h$ and $hu$ is given in Table \ref{tab:predict_swe_future}.

\begin{figure}[h!]
    \centering
    \includegraphics[width=0.3\linewidth]{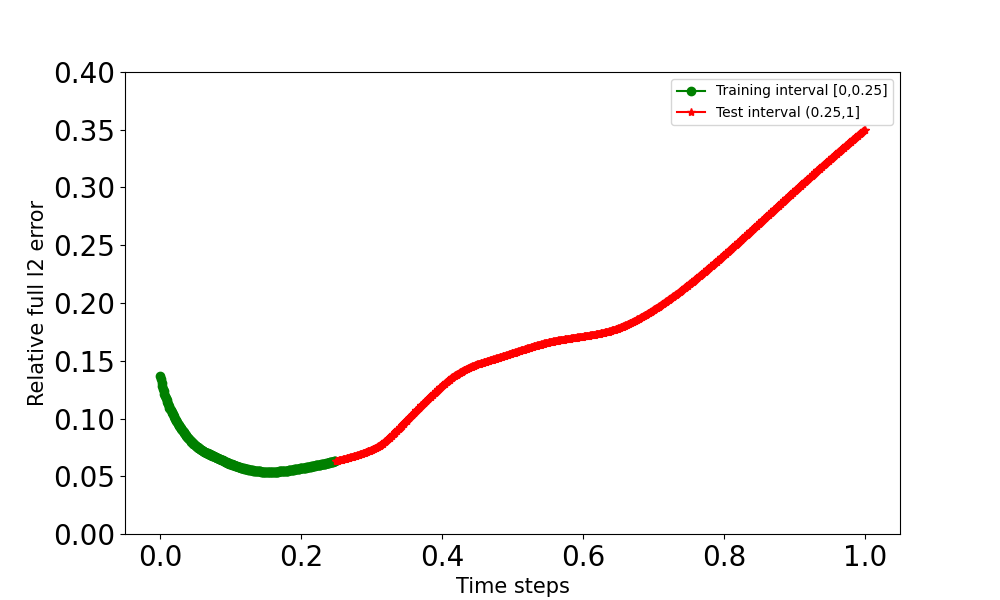}
    \includegraphics[width=0.3\linewidth]{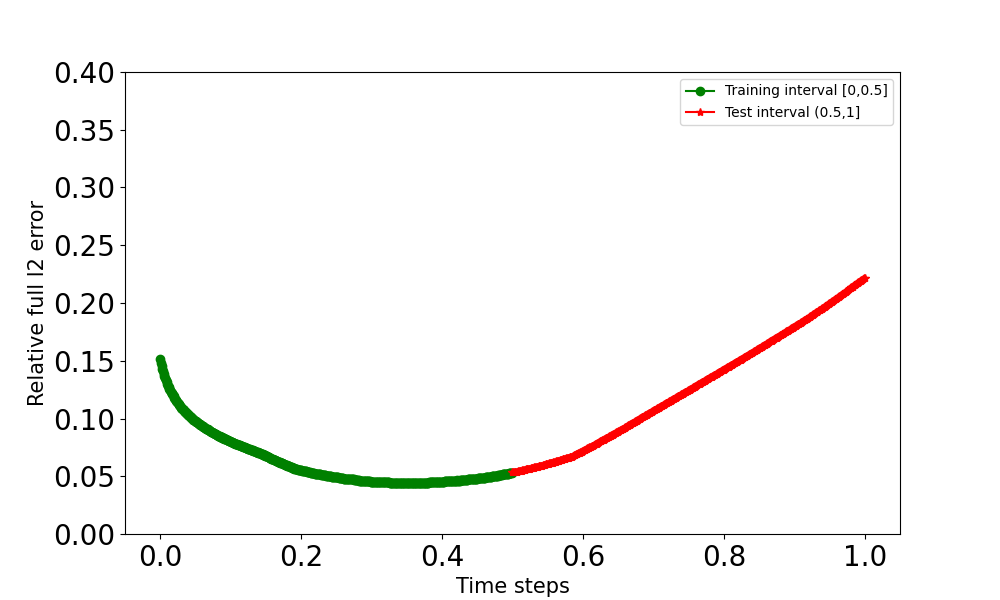}
    \includegraphics[width=0.3\linewidth]{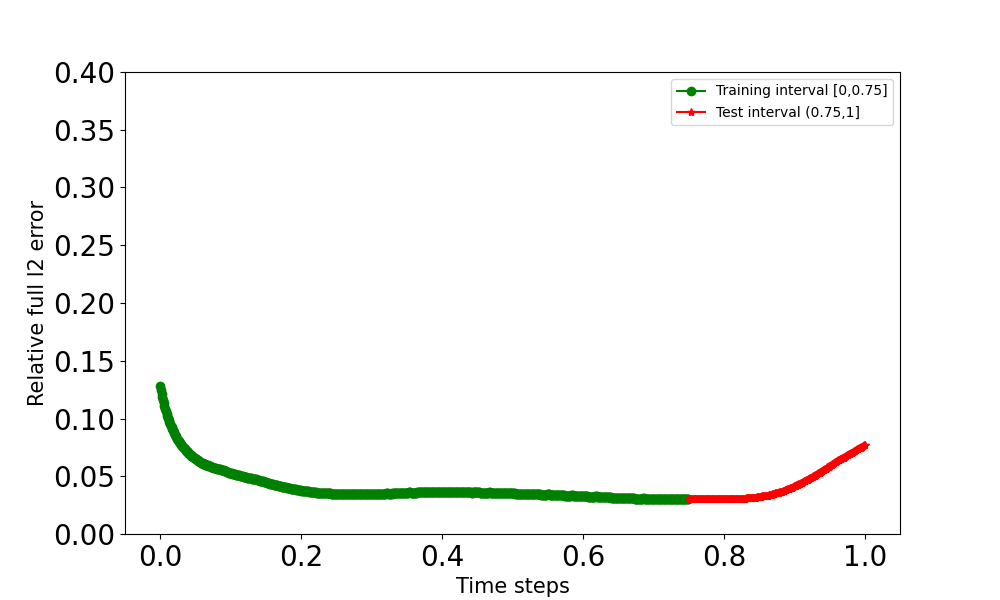}
    \caption{Error evolution of predicting the solution on an unseen temporal domain. Given in green is the full solution relative error for the time domain on which it is trained. In red, we have the error for the prediction domain.}
    \label{fig:error-forecast}
\end{figure}

\begin{table}[h!]
    \centering
    \begin{tabular}{|c|c|c|c|c|c|}
    \hline
         $N_t$ & $[t_0,t_1]$ & $(t_1,t_2] $ & RE for $h$ & RE for $hu$ & Mean RE of $[h,hu]^T$  \\
         \hline
         250 & [0,0.25] & (0.25,1] & 13.02\% $\pm$ 2.55 \% & 26.82 \% $\pm$ 7.59\% & 20.41\% $\pm$ 5.09\%\\
         500 & [0,0.5] & (0.5,1] & 7.53\% $\pm$ 5.99\% & 14.58\% $\pm$ 14.09\% & 11.68\% $\pm$ 10.96 \%\\
         750 & [0,0.75] & (0.75,1] & 3.22\% $\pm$ 0.54\% & 4.95\% $\pm$ 1.13\% & 4.26\% $\pm$ 0.80 \%\\
         \hline
    \end{tabular}
    \caption{Relative error (RE) for predicting solutions at unseen temporal points.}
    \label{tab:predict_swe_future}
\end{table}
We see that when the training time interval is small, the solution accuracy significantly drops. 
That is because the model only gets to learn the dynamics based on $[t_0,t_1]$ and is generalizing for a much larger domain $(t_1,t_2]$.
However, as we increase the training domain interval length $[t_0,t_1]$, the model predictions improve with training from the domain [0,0.75] giving an accuracy similar to that of using full domain data. 
This behavior is not unexpected since the model needs to see the solution evolution for enough number of timesteps to be able to generalize well. 
We have also given the plots of how the relative $\ell_2$ error for $[h,hu]^T$ evolves for the training time interval and the prediction interval in Figure \ref{fig:error-forecast}.
This gives us an insight into the number timesteps for which the OOD prediction errors will remain low before growing exponentially. 
The errors start to grow once when the prediction interval becomes far from the training domain.
Thus, in order to obtain an accurate forecast, the model must be trained with data that is closer to the forecasting time domain.
% Another aspect 
\subsection{Interchanging Data and Equation Variable}\label{sec:switched-data-eq-var}
In all the results above we selected a data variable and equation variable based on the complexity of the equation. 
The solution with more complex vector field was used as the data variable, assuming that a complex equation will be much harder to model and solve.
The one with less complex vector field was used as a physical prior.
In the case where both equations had similar complexity, we selected the data and equation variable randomly (in the RD system).
However, in reality it may be possible that data for the easier equation itself is available and a more complex equation is known to us.
We claim that even if we swap the data variable and equation variable, MUSIC would perform comparably. 
To demonstrate this, we swap the equation and data variable for the SWE system.
Now $h$ is used as data variable and $hu$ as the equation variable. Thus, the modified loss function is now defined as
\begin{equation}
\begin{aligned}
    \text{Loss} &= \dfrac{1}{N_sN_t}\sum\limits_{i=1}^{N_s}\sum\limits_{k=1}^{N_t}
    \left(\left\|h_{ik} - \widehat{h}_{ik}\right\|_2^2 +  \left\| ((\widehat{hu})_t)_{ik} + ((\widehat{h.}\widehat{u}^2)_x){_{ik}}+ \dfrac{1}{2}(g\widehat{h}^2)_x\right\|_2^2\right)\\ & + \dfrac{1}{N_{ic}}\sum\limits_{i=1}^{N_{ic}}\left\|(hu)_i - (\widehat{hu})_i\right\|_2^2 + \lambda_0\sum\limits_{j=1}^{|\text{Neurons}|}\sigma(\alpha_j).
    \end{aligned}
\end{equation}
Note that now the above loss uses data fitting loss for $h$ and physical constraints for $hu$.
We use the same setup as before i.e., we normalize the data using a min-max normalization and use a 80-20 training and validation split to train the model. 
We use the ADAM optimizer with learning rates tuned between $10^{-2}$ to $10^{-4}$.
\begin{table}[h!]
\centering
\small
\begin{tabular}{|c|c|c|c|}
\hline
$N_s$& $h$ & $hu$ & Full-field \\
\hline
50  & 0.011 \% & 0.061 \%& \(3.07\%\)\\
\hline
100  & 0.016\% & 0.067 \%& 3.65\% \\
\hline

\end{tabular}
\caption{Relative $\ell_2$ error of SWE model solutions learned with swapped priors by varying $N_s$. The model was trained using $N_t=800$, 4 Layers and 50 neurons. Final number of nonzero parameters was 3.26\% for both the cases.}
\label{tab:swe_swapped}
\end{table}
Since the model now learns using a different set of data and physical constraints, we modified the architecture of the model to a sinusodal activation function. 
Note that the solution $hu$ (see Figure \ref{fig:swe_true}) exhibits a wave-like behavior, which can be better captured by $\sin(\cdot)$ than a ReLU function.
Due to differing complexity of the loss function, the model hyperparameters (such as number of layers, hidden dimension, $N_s$, $N_t$, etc) giving the best trained model may be different from the ones used in Section \ref{sec:swe}.
Table \ref{tab:swe_swapped} shows that the model performance with swapped priors is similar to that of the main results described in Section \ref{sec:swe}.
Similarly, models using swapped priors for the other systems can also be trained.

\subsection{Limitations}
While the work tackles an important aspect of learning solutions from disjoint priors we acknowledge several limitations which serve as a basis for future work.
The amount of data required to obtain the depicted accuracy is comparatively higher than previous works that use full dimensional data \citep{chen2021physics}.
Since the work majorly focuses on learning the solutions, it needs to know the exact equation for one of the variables. 
If any term in the equation is unknown or inaccurately modeled, the accuracy of the learned solutions could be compromised. 
While for most cases neuron sparsity is useful, for other cases such as the wildfire system, thresholding algorithms can make the model training slower.
\section{Comparison to Prior Works}

Our proposed model is based on the assumption that available information i.e., the knowledge of the physical equation of a given variable, and the given data are disjoint. 
While there are no existing works in literature which specifically work with learning solutions of a coupled system with disjoint priors, the loss function defined in MUSIC is similar to the approach considered in \citep{raissi2019physics} where the major difference is that they assume full system knowledge or data availability for all the variables.
Some other applications of PINNs using incomplete data can be found in \citep{xu2023practical,kuo2023physics}.
This further undermines the efforts presented in this paper where we demonstrate the possibility of learning the full solution of a coupled PDE when the priors are fully disjoint, with no overlap of information.
Moreover, our work incorporates model sparsity to tackle sparse information and data in order to learn the solutions. 
Previous works in \citep{chen2021physics} have explored the possibility of learning the system equations using sparsity, similar to SINDy \citep{brunton2016discovering}, however they do not consider model sparsity.
We include a comparison of our proposed model with PINN \citep{raissi2019physics} and PINN-SR \citep{chen2021physics} in Table \ref{tab:ours-vs-pinn}.
\begin{table}[h!]
    \centering
    % \small
    \begin{tabular}{|c|p{3cm}|p{2cm}|p{3cm}|}
    \hline
        Name & MUSIC & PINN \citep{raissi2019physics} & PINN-SR \citep{chen2021physics} \\
        \hline
        % Method & Physical constraint and model $\ell_0$ sparsity. & Physical constraints with $\ell_2$ loss. & Physical constraints with sparse dictionary learning. \\
        % \hline
        Sparsity type & $\ell_0$ & None & $\ell_0$\\
        \hline
        Sparse components & Model neurons or weights & None & Dictionary (coefficient) weights\\
        % \hline
         % Model sparsity & $\checkmark$ & $\times$ & $\times$\\
        \hline
        Sparse learning & $\checkmark$ & $\times$ & $\checkmark$\\
        \hline
         Model compression & $\checkmark$ & $\times$ & $\times$\\ 
        % \hline
        % Training Data & Low & High & Low \\
        \hline
        Disjoint Priors & $\checkmark$ & $\times$ & $\times$ \\
        \hline    
        Collocation Points & $\times$ & $\checkmark$ & $\checkmark$ \\
        \hline
    \end{tabular}
    \caption{Comparison of our work with similar prior works.}
    \label{tab:ours-vs-pinn}
\end{table}
In order to have a fair comparison of our proposed model against prior works, we consider two other benchmark models: (a) model without any sparsity; (b) two decoupled models results where one neural network learns the solutions from data (function approximation using supervised learning) and the another neural network learns the solutions from equations using PINNs. 

Suppose, $u_1, u_2$ denotes the full-field solution of a coupled system, where scarce data for $u_1$ is given and the equation $\dfrac{\partial u_2}{\partial t} = F_2(u_1,u_2)$ is known. We describe the model structure and loss for our comparison methods below. 
% \subsection{Setup for Benchmark Methods}
\begin{enumerate}
    \item \textbf{PIML$_{inc}$}: Short for incomplete Physics Informed Machine Learning, in this method we train a neural network model without considering any kind of $\ell_0$ regularization. The loss function is given as
    \begin{equation}
        L = \lambda_1\|u_1 - \widehat{u}_1\|_2^2 + \lambda_2\left\|\dfrac{\partial\widehat{u}_2}{\partial t} - F_2(\widehat{u}_1,\widehat{u}_2)\right\|_2^2 + \lambda_{ic}\|u_{2_{ic}} - \widehat{u}_{2_{ic}}\|_2^2
    \end{equation}
    where $\widehat{u}_1$, $\widehat{u}_2$ denote the outputs of a multitask neural network and $u_{2_{ic}}$ denotes data at the initial time.
    As the name suggests, the model is very similar to a full PINN model proposed in \citep{raissi2019physics} with only the terms for which the information is available included in the loss function. 
    Our main comparison is with this model since it is very similar to the proposed structure, however differing in sparsity. We compare all the simulated examples above with this method.
    
    \item \textbf{NN$_{PIML+Data}$}: For this method, we consider a direct approach where we treat both $u_1$ and $u_2$ as decoupled variables and learn each solution using a separate NN model. Suppose $NN_1$ and $NN_2$ denotes the neural networks used to learn solutions $u_1$ and $u_2$. We learn $u_1$ using
    \begin{equation}
    \begin{aligned}
        \widehat{u}_1 &= NN_1(\mathbf{x},t)\\ 
        L_1 &= \|u_1 - \widehat{u}_1\|_2^2.
    \end{aligned}
    \end{equation}
    The solutions $u_2$ is learned using 
    \begin{equation}
    \begin{aligned}
        \widehat{u}_2 &= NN_2(\mathbf{x},t)\\ 
        L_2 &= \lambda_2\left\|\dfrac{\partial\widehat{u}_2}{\partial t} - F_2(u_1,\widehat{u}_2)\right\|_2^2 + \lambda_{ic}\|u_{2_{ic}} - \widehat{u}_{2_{ic}}\|_2^2.
    \end{aligned}
    \end{equation}
    We directly use the data available for $u_2$ as an input in $L_2$. 
    % We use this method to compare with the overall best performing model for 
\end{enumerate}
% We compare our method with the above mentioned work(s) in detail for two systems: the SWE system that has only one spatial dimension; and the FN system that has two spatial dimensions. 
% \subsubsection*{Shallow Water Equations}
\subsection*{Neurons vs layers}
\begin{figure}[ht!]
     \centering
     \begin{subfigure}[b]{0.49\textwidth}
         \centering
         \includegraphics[width=1.0\linewidth]{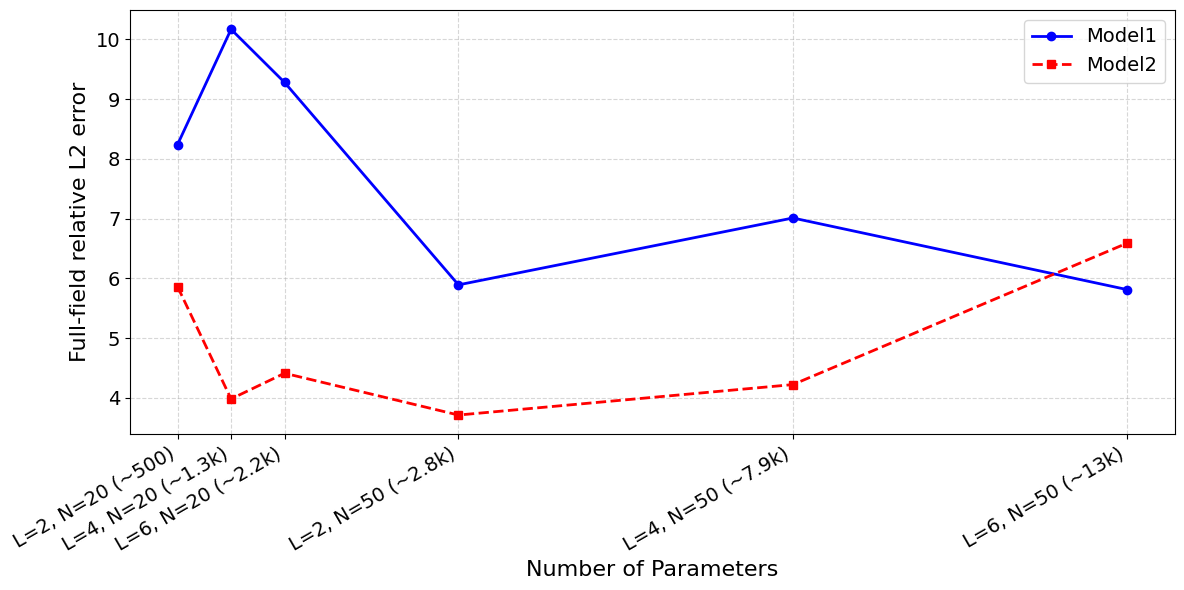}
    \caption{SWE System. $N_s = 100$, $N_t = 800$.}
    \label{fig:swe-compare-neurons}
     \end{subfigure}
        \begin{subfigure}[b]{0.49\textwidth}
         \centering
         \includegraphics[width=1.0\linewidth]{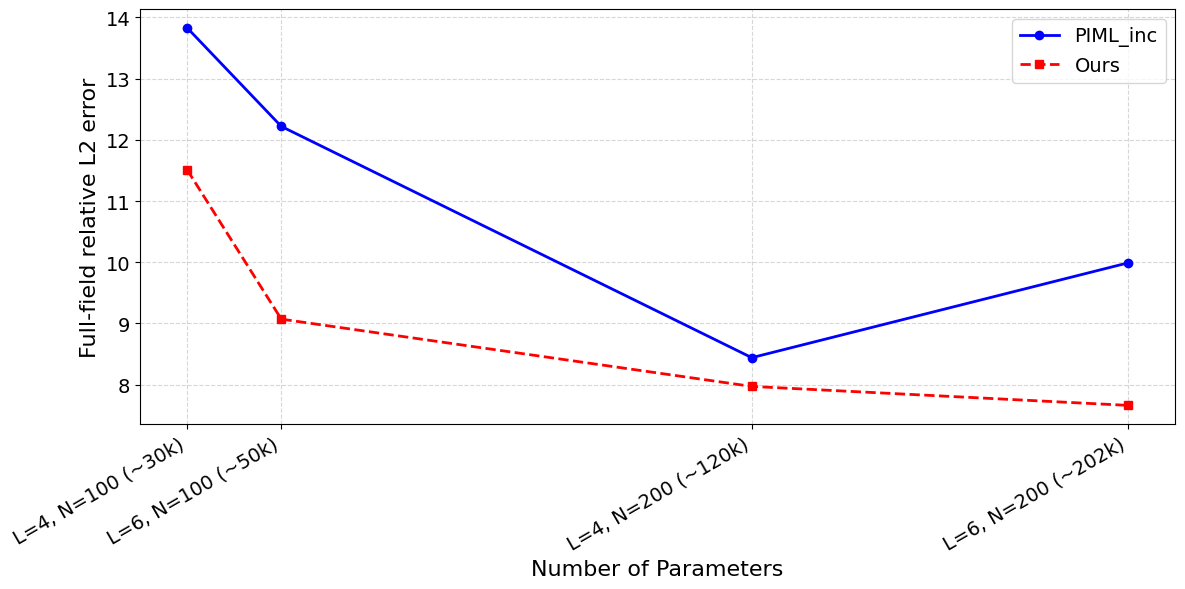}
    \caption{FN system. $N_s = 5000$, $N_t = 100$.}
    \label{fig:fn-compare-neuron-layers}
     \end{subfigure}
         \begin{subfigure}[b]{0.49\textwidth}
         \centering
          \includegraphics[width=1.0\linewidth]{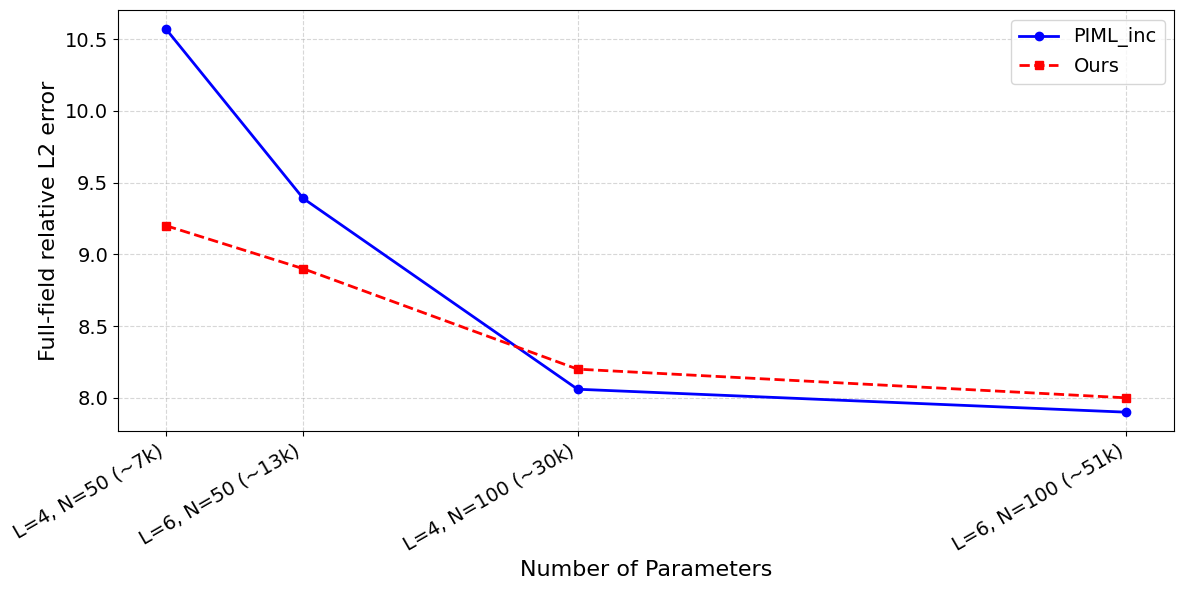}
    \caption{RD system. $N_s = 5000$, $N_t = 100$.}
    \label{fig:rd-compare-neurons}
     \end{subfigure}
         \begin{subfigure}[b]{0.49\textwidth}
         \centering
         \includegraphics[width=1.0\linewidth]{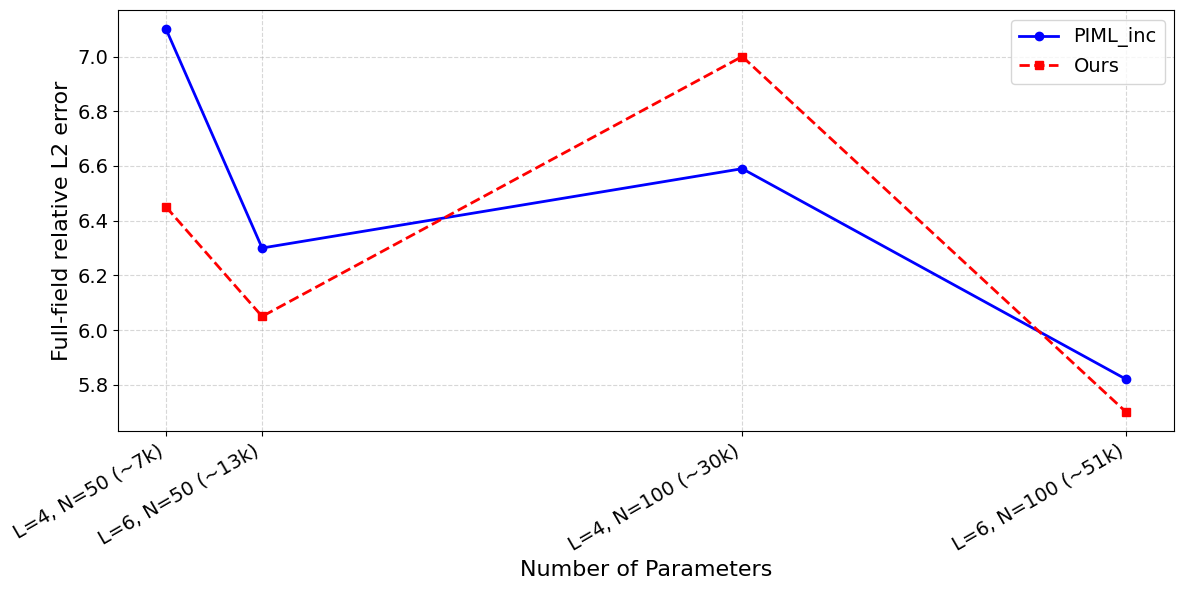}
         \caption{Wildfire System. $N_s=1000$, $N_t=60$.}
         \label{fig:wildfire-compare-neurons}
     \end{subfigure}
        \caption{Comparison of full-field relative $\ell_2$ error (in percentage) for different layers and neurons for MUSIC and PIML$_{inc}$.}
        \label{fig:neuron-layers-compare-full}
\end{figure}

From Figure \ref{fig:neuron-layers-compare-full}, we plot the full-field average $\ell_2$ error for the proposed model with the PIML$_{inc}$ for different layers and hidden dimension. 
The plots for SWE, FN, RD and Wildfire systems are given in Figures \ref{fig:swe-compare-neurons}, \ref{fig:fn-compare-neuron-layers}, \ref{fig:rd-compare-neurons} and \ref{fig:wildfire-compare-neurons}, respectively. 
The number of spatial data $N_s$ and $N_t$ for each method is specified in the caption. Our model with $\ell_0$ regularization outperforms the PIML$_{inc}$ model for majority of the cases.
Especially for lesser complex models (for example, with 4 layers only), the $\ell_0$ regularization helps the model to get lower accuracy for lesser complex models.
The advantage of using regularization is more evident for systems whose corresponding equations for variables $u_1$ and $u_2$ are complex and different from each other.
For example, in SWE system the equations for $h$ and $hu$ are very different from each other. Similarly for the FN system, the equations for $u$ and $v$ look very different.
Thus we see in Figures \ref{fig:swe-compare-neurons} and \ref{fig:fn-compare-neuron-layers}, regularization plays a stronger role in improving the results.
For the RD system, both equations for $u$ and $v$ are very similar, leading to both the models performing similarly with negligible differences.
The wildfire system, due to its discontinuity and dominance of constant values in the training domain uses the thresholding approach for sparsity constraints. 
Since the thresholding mainly induces unstructured sparsity, the main advantage can be seen in model pruning, with both PIML$_{inc}$ and our proposed model performing comparably.

% \subsubsection*{FN System of Equations}
\subsection*{Model Learning from Noisy Data}
In this section, we compare MUSIC with PIML$_{inc}$ for the systems where we assessed model learning using noisy data.
Given in Table \ref{tab:errors-noise-full} are the full-field average relative errors of our proposed model with PIML$_{inc}$ with \textcolor{blue}{$\downarrow$} indicating an lower error and \textcolor{red}{$\uparrow$} indicating higher error in comparison to PIML$_{inc}$. 
For a majority of the cases, we see that when the noise is small (2\% for SWE and FN, and 5\% for RD), the improvement is minor, often below 1\% for majority of the cases, except for FN system with $N_{s_x}*N_{s_y}=500$ where the error improves by more than 2\%. 
However, as we increase the noise level to 10\%, the error falls by about 1.5\% (more or less) except for the FN case with $N_{s_x}*N_{s_y}=500$ where the error marginally goes up.
For the RD system, where both $u$ and $v$ have very similar equations, for lower noise levels, the errors are very similar, a trend observed even for the previous case (Figure \ref{fig:rd-compare-neurons}) with clean data.
However, the regularization plays an important role when the data contains 20\% noise and very less spatial training data of $N_{s_x}*N_{s_y}=500$, where we see a 2.35\% reduced error from PIML$_{inc}$.
\begin{table}[h!]
    \centering
    \small
    \resizebox{\textwidth}{!}{%
    \begin{tabular}{|c|c|c|c|c|cc|cc|cc|}
        \hline
        \multirow{1}{*}{System} & \multirow{1}{*}{L} & \multirow{1}{*}{H} & \multirow{1}{*}{$N_t$} & \multirow{1}{*}{$N_{s_x}*N_{s_y}$} 
        & MUSIC & PIML$_{inc}$ & MUSIC & PIML$_{inc}$ & MUSIC & PIML$_{inc}$  \\
        % \cline{6-11}
        \hline
         & & & & & \multicolumn{2}{c|}{2\% Noise} & \multicolumn{2}{c|}{5\% Noise} & \multicolumn{2}{c|}{10\% Noise}  \\
          % & & & & & MUSIC & PIML & MUSIC & PIML & MUSIC & PIML \\
        \cline{6-11}
        \multirow{3}{*}{SWE} & \multirow{3}{*}{4} & \multirow{3}{*}{20} & \multirow{3}{*}{800} &50 & 4.80 (\textcolor{red}{$\uparrow$}0.04\%) & 4.76 & 4.77 (\textcolor{blue}{$\downarrow$}1.01\%) & 5.78 & 5.29 (\textcolor{blue}{$\downarrow$}2.44\%) & 7.44 \\
         &   &    & &100 & 4.83 (\textcolor{blue}{$\downarrow$}0.09\%) & 4.92 & 5.07 (\textcolor{blue}{$\downarrow$}0.86\%) & 5.93 & 5.62 (\textcolor{blue}{$\downarrow$}1.37\%) & 6.99 \\
         &   &    & &150 & 4.27 (\textcolor{blue}{$\downarrow$}0.94\%) & 5.21 & 4.50 (\textcolor{blue}{$\downarrow$}0.85\%) & 5.35 & 6.27 (\textcolor{blue}{$\downarrow$}2.32\%) & 8.59 \\
        \hline
        & & & & & \multicolumn{2}{c|}{2\% Noise} & \multicolumn{2}{c|}{5\% Noise} & \multicolumn{2}{c|}{10\% Noise}  \\
        \cline{6-11}
        \multirow{2}{*}{FN} & \multirow{2}{*}{6} & \multirow{2}{*}{100} & \multirow{2}{*}{40} &500 & 16.28 (\textcolor{blue}{$\downarrow$}2.4\%) & 18.68 & 21.88 (\textcolor{blue}{$\downarrow$}0.19\%) & 22.07 & 27.98 (\textcolor{red}{$\uparrow$}0.95\%) & 27.03 \\
         &   &     & & 1000 & 14.98 (\textcolor{blue}{$\downarrow$}0.78\%) & 15.76 & 22.39 (\textcolor{red}{$\uparrow$}2.28\%) & 20.11 & 24.87 (\textcolor{blue}{$\downarrow$}1.41\%) & 26.28 \\
        \hline
        & & & & & \multicolumn{2}{c|}{5\% Noise} & \multicolumn{2}{c|}{10\% Noise} & \multicolumn{2}{c|}{20\% Noise}  \\
        \cline{6-11}
        \multirow{2}{*}{RD} & \multirow{2}{*}{6} &\multirow{2}{*}{100} & \multirow{2}{*}{100} &500 & 11.34 (\textcolor{blue}{$\downarrow$}0.01\%) & 11.35 & 10.94 (-) & 10.94 & 11.98 (\textcolor{blue}{$\downarrow$}2.35\%) & 14.33 \\
         &   &     & & 1000    & 8.55 (\textcolor{blue}{$\downarrow$}0.21\%) & 8.75  & 9.29 (\textcolor{red}{$\uparrow$} 0.54\%) & 8.42 & 10.94 \textcolor{blue}{$\downarrow$}0.28\%) & 11.22 \\
        \hline
    \end{tabular}
    }
    \caption{Relative errors of different models under varying noise levels and sample sizes. Note ofr SWE, $N_{s_x}*N_{s_y} = N_s$.}
    \label{tab:errors-noise-full}
\end{table}
% \begin{figure}[h!]
%     \centering
%     \includegraphics[width=0.5\linewidth]{FHN/fhn-neur-layer-compare.png}
%     \caption{Comparison of full-field relative $\ell_2$ error (in percentage) for different layers and neurons for MUSIC and PIML$_{inc}$. Both models were trained using $N_s = 5000$ and $N_t = 100$.}
%     \label{fig:fn-compare-neuron-layers}
% \end{figure}

% \begin{table}[h!]
% \small
%     \centering
%     \begin{tabular}{|c|c|c|c|c|c|c|c|c|c|c|}
%        \hline
%       \backslashbox{$N_s$}{Noise}  & \multicolumn{3}{c|}{2\%} & \multicolumn{3}{c|}{5\%} & \multicolumn{3}{c|}{10\%}\\
%       \hline
%       Model & MUSIC & PIML & NN& MUSIC & PIML & NN & MUSIC & PIML & NN \\
%       \hline
%       % \hline
%        500  &  \textbf{16.28} & 18.68 & 19.24 & \textbf{21.88} & 22.07 & 27.21 & 27.98 & \textbf{27.03} & 32.49\\
%        \hline
%        1000  & \textbf{14.98} &15.76&19.93 & 22.39 &\textbf{20.11} &29 & \textbf{24.87} & 26.28 & 33.25\\
%        \hline
%     \end{tabular}
%     \caption{Comparison of the full-field average relative $\ell_2$ error (in percentage) for different noise levels. With $N_t = 800$, based on multiple simulations, the best performing model for each type was used. MUSIC results are based on 4 layers and 20 neurons while PIML$_{inc}$ uses 4 layers and 100 neurons. \textbf{For the sake of formatting PIML$_{inc}$ and NN$_{PIML+Data}$ is abbreviated as PIML and NN respectively.} }
%     \label{tab:fhn-compare-noise}
% \end{table}
\subsubsection*{Out of Domain Forecasting}
In this section, we briefly discuss the performance of PIML$_{inc}$ with respect to predicting for out of domain temporal points i.e., when the forecasting interval is disjoint from the training interval. 
Referring to Section \ref{sec:swe-forecast}, we analyzed how MUSIC forecasts the solution $[h,hu]^T$ of the SWE system when it is trained on $[t_1,t_2]$ and tested on $(t_2,t_3]$ such that $[t_1,t_2]\cap (t_2,t_3]=\emptyset$.
In Table \ref{tab:swe-compare-forecast} we report the errors for forecasting $h$ and $hu$ when different $t_1,t_2,t_3$ are used.
The reported errors are for the proposed model. In brackets given are the amount by which MUSIC's error differ from PIML$_{inc}$.
Red arrows denote an increase in error and blue arrows denote a decrease in error in comparison to PIML$_{inc}$.
A significant improvement can be noted when the training interval is strictly smaller than the forecasting interval. A significant improvement of 11\% can be found when learning from data under extremely data scarce conditions. Overall there is a 4.36\% improvement.
Overall, the model improves significantly when regularization is used.

\begin{table}[h!]
    \centering
    \begin{tabular}{|c|c|c|c|c|}
    \hline
          Train & Forecast & Error $h$ & Error $hu$ & Full-field error  \\
    \hline
          [0,0.25] & (0.25,1] & 13.02 (\textcolor{red}{\textcolor{red}{$\uparrow$}} 1.3\%) & 26.82 (\textcolor{blue}{\textcolor{blue}{$\downarrow$}} 11\%) & 20.41 (\textcolor{blue}{\textcolor{blue}{$\downarrow$}} 4.36\%)\\
    \hline
          [0.25,0.5] & (0.5,1] & 7.53 (\textcolor{red}{\textcolor{red}{$\uparrow$}} 0.14\%) & 14.58 (\textcolor{blue}{\textcolor{blue}{$\downarrow$}} 3.16\%) & 11.68 (\textcolor{blue}{\textcolor{blue}{$\downarrow$}} 0.89\%) \\
    \hline
          [0,0.75] & (0.75,1] & 3.22 (\textcolor{blue}{\textcolor{blue}{$\downarrow$}} 1.95\%) &4.95 (\textcolor{blue}{\textcolor{blue}{$\downarrow$}} 3.65\%) & 4.26 (\textcolor{blue}{\textcolor{blue}{$\downarrow$}} 2.62\%)\\
    \hline
    \end{tabular}
    \caption{\small Relative error (in percentage) for predicting solutions at unseen temporal points. Arrows show error difference from a model without regularization (PIML$_{inc}$).}
    \label{tab:swe-compare-forecast}
\end{table}

% \subsubsection*{RD system}
\subsection*{Full comparison}
For the sake of completeness of the paper, we include a brief summary of the comparison of the best results obtained using various models.
In Table \ref{tab:all-compare-best}, we compare the the best trained models for each of the systems which include both PIML$_{inc}$ and NN$_{PIML+Data}$ using clean data.
We report the relative $\ell_2$ error (in percentage) for the data variable, equation variable as well as the full-field solution.
The lowest error for each noise level and $N_s$ is highlighted. 
For four out of the 6 cases, MUSIC outperforms all the models with the PIML$_{inc}$ method outperforming MUSIC marginally for the other two cases. 
We see that using two separate networks does not perform better for any of the cases indicating that NN$_{PIML+Data}$ method fails to capture any dependencies between the two interactive variables.
\begin{table}[h!]
    \centering
    \resizebox{\textwidth}{!}{%
    \begin{tabular}{|c|c|c|c|c|c|c|c|c|c|}
    \hline
         System &\multicolumn{3}{c|}{Data} & \multicolumn{3}{c|}{Equation} & \multicolumn{3}{c|}{Total}  \\
         \hline
         & MUSIC & PIML & NN & MUSIC & PIML & NN &MUSIC & PIML & NN \\
         \hline
         SWE & 3.07\% & 2.48\% & 2.15\% & 7.93\% & 6.84\% & 21.69 & 3.98\% & 4.56\% & 12.19 \\
         FN & 3.5\% & 5.21\% & 10.67\% & 10.96\% & 9.55\% & 11.38\% & 7.64\% & 8.44\% & 11.025\% \\
         RD & 4.6\% & 4.3\% & 3.69\% & 10.26\% & 10.31\% & 11.36\% & 7.65\% & 7.9\% & 7.52\% \\
         Wildfire & 8.14\% & 8.79 & 8.33\% & 5.67\% & 5.48 & 4.87\% & 5.8\% & 5.96\% & 6.44\%\\
         \hline  
    \end{tabular}
    }
    \caption{Comparison of the best trained models for PIML$_{inc}$ (PIML in the table) and NN$_{PIML+Data}$ (NN in the table) with MUSIC.}
    \label{tab:all-compare-best}
\end{table}
We train all the models using layer $L\in\{4,6\}$ and hidden dimension $H\in\{50,100\}$ and report the errors for the best performing model for each of the methods. 
The lowest errors for the data variable in many cases is given by the NN model. 
The NN model learns the solutions in a decoupled way, leading to higher errors in general. 
When we compute the full-field relative $\ell_2$ error, we see the usefulness of training a multitask model where both MUSIC and PIML give lower errors in majority of the cases.
Moreover, the MUSIC is capable of model compression since we induce sparsity. 
The trained models using MUSIC have atleast 90\% sparsity (zero weights) in comparison to PIML$_{inc}$ when using structured sparsity, and atleast 50\% sparsity when using unstructured sparsity.
However, the other models are dense and hence would require significantly more resources to store. 
Also note that for the NN model, double the computational resources are required since it trains two models separately.

% \begin{figure}
%     \centering
%     \includegraphics[width=0.5\linewidth]{rd-spiral/rd-neuron-vs-layers-compare.png}
%     \caption{Comparison of full-field relative $\ell_2$ error (in percentage) for different layers and neurons for MUSIC and PIML$_{inc}$. Both models were trained using $N_s = 5000$ and $N_t = 100$.}
%     \label{fig:placeholder}
% \end{figure}

% \begin{table}[h!]
%     \centering
%     \begin{tabular}{|c|c|c|c|c|c|c|c|}
%     \hline
%           Train dom. & Forecast dom. & \multicolumn{2}{c|}{Error $h$} & \multicolumn{2}{c|}{Error $hu$} & \multicolumn{2}{c|}{Full-field error}  \\
%          \hline
%          $[t_0,t_1]$&$(t_1,t_2] $ & MUSIC & PIML$_{inc}$ &MUSIC & PIML$_{inc}$ &MUSIC & PIML$_{inc}$ \\
%          \hline
%           [0,0.25] & (0.25,1] & 13.02 (\textcolor{red}{$\uparrow$} 1.3\%) &11.72 &26.82 (\textcolor{blue}{$\downarrow$} 11\%) &37.82 & 20.41 (\textcolor{blue}{$\downarrow$} 4.36\%) & 24.77\\
%           % [0,0.5] & (0.5,1] & 7.53 &7.39 & 14.58 & 17.74 & 11.68 & 12.57\\
%           % [0,0.75] & (0.75,1] & 3.22&5.17 &4.95 &8.60 & 4.26&6.88\\
%          % \hline
%     \end{tabular}
%     \caption{Relative error for predicting solutions at unseen temporal points.}
%     \label{tab:predict_swe_future}
% \end{table}

% \section{Applications using Real Data for Wildfire Spread}
% \subsection{Wildfire Spread in Jasper in July 2024}
\section{Conclusion}
In this paper, we propose a regularized approach for solving coupled PDEs when the available data and physical prior is mutually exclusive.
The proposed problem statement is inspired from real world applications where we often have data for one variable and some known physics for another variable and there is no way to learn them since there is no overlap of data (or physical knowledge) between these two coupled variables. 
We use a physics informed approach along with $\ell_0$ regularization to induce model compression and tackle the issue of data sparsity.
The proposed approach is tested on four benchmark systems commonly found in literature, the SWE system, FN system, RD system and a wildfire model (special case of RD system). 
For each of the systems, we discuss in detail the performance of the proposed approach with respect to available data, noisy training data, varying model complexity (layers versus hidden dimension), model compression, etc.
Although we did not find any direct method addressing disjoint priors in literature, we compared our method by constraining the loss functions of the other well known approaches such as PINN and NN.
Our proposed approach not only outperformed the other methods for majority of the cases, it achieved a much higher level of model compression leading to better generalizability for lower levels of training data, out of domain forecasting and noisy datasets.

\section*{Code and Data Availability}
All codes and data used to generate the results can be accessed at \url{https://github.com/esha-saha/coupled-system-solutions-with-disjoint-priors.git}.

\section*{Acknowledgements}
E.S. was funded by the Grant Notley Memorial Postdoctoral Fellowship. The authors would like to thank Dr. Giang Tran for her insightful comments. 

\bibliographystyle{abbrv}
\bibliography{references}

\end{document}